\documentclass[mlabstract]{jmlr}

\usepackage{mathtools}
\DeclareMathOperator{\tr}{tr}
\DeclareMathOperator{\vectorize}{vec}
\DeclareMathOperator{\diag}{diag}
\DeclareMathOperator{\var}{var}
\DeclareMathOperator{\image}{im}
\DeclarePairedDelimiter{\norm}{\lVert}{\rVert}

\usepackage{longtable}

\usepackage{booktabs}
\usepackage[load-configurations=version-1]{siunitx} 

\theorembodyfont{\upshape}
\theoremheaderfont{\scshape}
\theorempostheader{:}
\theoremsep{\newline}

\jmlrvolume{}
\firstpageno{1}
\editors{List of editors' names}

\jmlryear{2026}
\jmlrworkshop{Symmetry and Geometry in Neural Representations}

\title{Broken scale symmetries in \texorpdfstring{ \\}{} undercomplete linear autoencoders}

 \author{\Name{Farhad Pashakhanloo} \Email{fpasha@fas.harvard.edu}\\
 \addr Center for Brain Science\\
      Harvard University
 \AND
 \Name{Jacob A. Zavatone-Veth} \Email{jzavatoneveth@fas.harvard.edu}\\
 \addr Center for Brain Science and Society of Fellows\\
      Harvard University
 }

\begin{document}

\maketitle

\begin{abstract}
Neural network loss landscapes have many symmetries, which are preserved by gradient flow but broken by finite-stepsize stochastic gradient descent (SGD). A canonical example of such a symmetry is scale in homogeneous networks: one can scale up the parameters in one layer and down in the next without changing the network output. Previous work has documented cases in which SGD breaks this symmetry in favor of balancing gradient noise or minimizing fluctuations. Here, we show that the solution geometry of undercomplete linear autoencoders instead selects a preferred sign for scale drift: on the PCA solution manifold, SGD favors large decoder weights. This directed scale drift occurs on a slow timescale, and its dynamics admit an analytically-tractable effective description. However, it cannot continue indefinitely: increasing scale eventually drives the dynamics towards a finite-stepsize stability boundary. The resulting solutions are sharper than a balanced baseline in the sense of the maximum eigenvalue of the loss Hessian, but different sharpness measures can move in opposing directions. Thus, undercomplete autoencoders give a concrete illustration of how loss geometry can convert residual gradient noise into directed motion along a manifold of  functionally-equivalent solutions.
\end{abstract}
\begin{keywords}
stochastic gradient descent, representation learning, parameter symmetries
\end{keywords}

\setlength{\abovedisplayskip}{0.1em}
\setlength{\belowdisplayskip}{0.1em}

\section{Introduction}

How does SGD navigate symmetry-induced flat directions in neural network loss landscapes? Previous works have proposed a range of perspectives on this inductive bias, from effective regularizers \citep{li2022what,ziyin2025thermodynamics} and equilibrium balance conditions \citep{ziyin2025noise,ziyin2025thermodynamics}, to manifold diffusion \citep{pashakhanloo2023drift} and sparsifying collapse \citep{chen2024collapse}. However, it remains unclear what determines the direction and endpoint of SGD motion along continuous symmetry orbits. Here, we study a solvable model---undercomplete linear autoencoders---in which we can give a precise answer to this question in the specific case of scale symmetries. We show that in this setting solution geometry selects a preferred direction for drift along scale-symmetry directions, and solve for its slow dynamics. 

\section{Undercomplete solution geometry results in directed scale drift}

We consider a tractable toy setting: linear autoencoders
\begin{align}
    f(x) = W U x,
\end{align}
with encoder weights $U \in \mathbb{R}^{k \times d}$ and decoder weights $W \in \mathbb{R}^{d \times k}$. We train these models to minimize an $L^2$ reconstruction loss $L = \frac{1}{2}\mathbb{E}_{x}[\Vert f(x) - x \Vert^2]$, assuming zero-mean data $x \in \mathbb{R}^{d}$ with covariance $\Sigma = \mathbb{E}[xx^{\top}]$. Our focus is on the undercomplete regime $k < d$, where exact reconstruction is impossible if $\Sigma$ is of full rank. 

\begin{figure}[t]
    \centering
    \hspace*{-0.1in}
    \includegraphics[width=\linewidth]{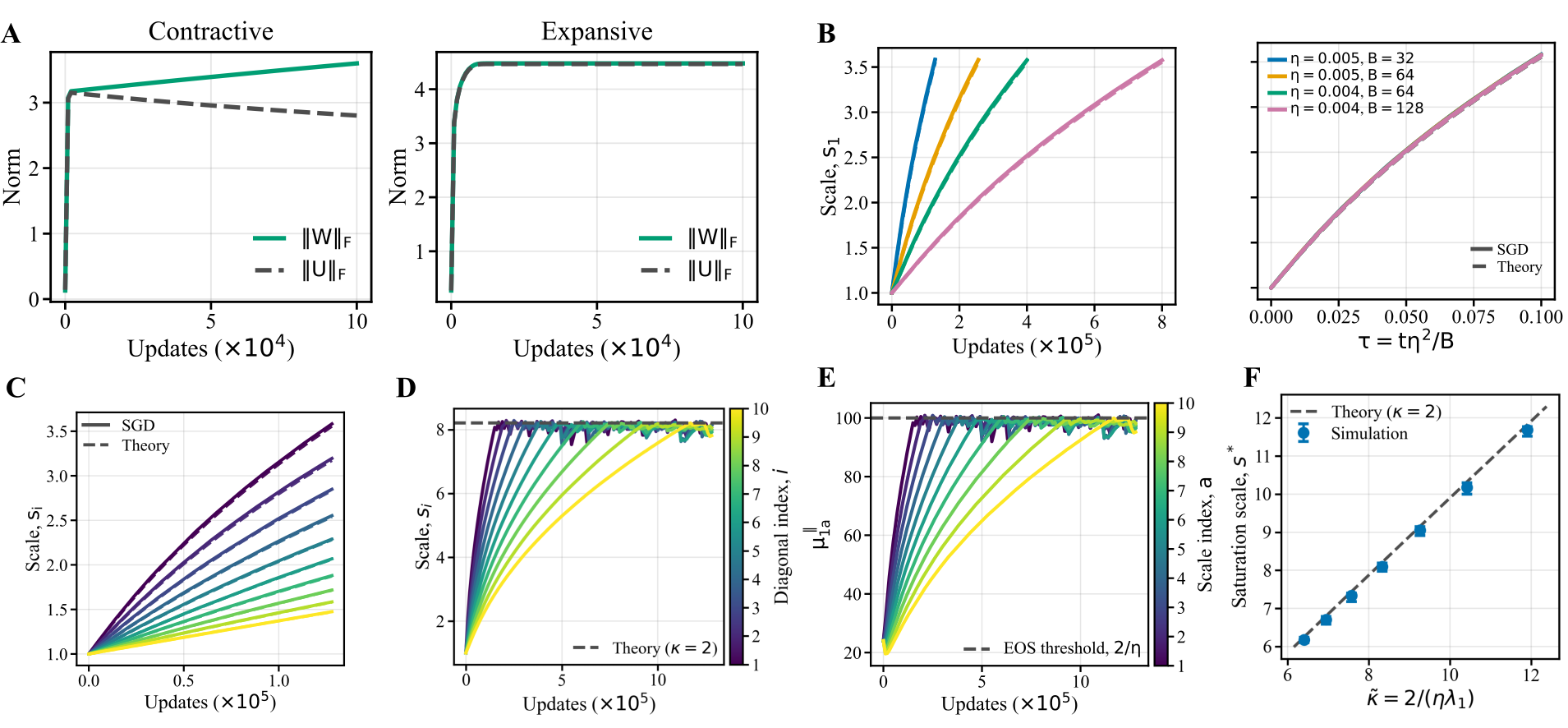}
    \vspace{-1em}
    \caption{(A)
    Continual minibatch training drives scale imbalance in linear autoencoders.
    Frobenius norms of decoder and encoder weights,
    $\|W\|_F$ and $\|U\|_F$, for contractive and expansive autoencoders, respectively (see App.~\ref{app:experiments} and Fig.~\ref{fig:linear-scale-imbalance} for additional details).
    (B) Scale growth in a contractive linear autoencoder for different learning rates ($\eta$) and minibatch size ($B$). $s_1 = (W^{\top}W)_{11}$ is the dominant scale. Theory predictions are based on Eq.~\ref{eqn:model_scale_growth_main}. 
    (C) Anisotropic scale growth in a contractive autoencoder. Different scale modes, $s_i = (W^{\top}W)_{ii}$, grow at different rates. Theory predictions are based on Eq.~\ref{eqn:model_scale_growth_main}. (D-F) Edge of Stability phenomenon driven by imbalancing dynamics. (D) Scale growth and saturation for different modes. (E) Sequential saturation of eigenvalues. (F) Saturation as a function of the inverse of the learning rate and comparison to theory.}
\label{fig2:workshop}
\vspace{-1em}
\end{figure}

These networks have a $\mathrm{GL}(k)$ symmetry group that preserves the function output: mapping $(W,U) \mapsto (W A, A^{-1} U)$ for $A \in \mathrm{GL}(k)$ leaves $f(x)$ invariant. Under gradient flow, it is well-known that the Noether charge 
\begin{align} \label{eqn:bigQ_definition}
    Q = W^{\top} W - UU^{\top}  \in \mathbb{R}^{k \times k}
\end{align}
corresponding to this symmetry is conserved, \textit{i.e.}, $\partial_{t} Q = 0$ \citep[App.~\ref{app:pca_basics};][]{tanaka2021noether,kunin2021neural}. $Q$ measures imbalances in the incoming and outgoing weights for each latent neuron. In contrast, under SGD with learning rate $\eta$ and loss $\hat{L}_{t}=\frac{1}{2}\mathbb{E}_{x\sim\mathcal{B}_{t}}[\norm{f(x)-x}^2]$ for a minibatch $\mathcal{B}_{t}$ of size $B$ and covariance $\hat{\Sigma}_{t}$, $Q$ evolves as
\begin{align} \label{eqn:bigQ_general_update}
    Q_{t+1} = Q_{t} + \eta^{2} [(\nabla_{W}\hat{L}_{t})^{\top}(\nabla_{W}\hat{L}_{t}) - (\nabla_{U}\hat{L}_{t})(\nabla_{U}\hat{L}_{t})^{\top}] 
\end{align}
Therefore, SGD will generically break the symmetry, as has been noted in general terms in prior works \citep{tanaka2021noether,kunin2021neural,ziyin2025noise}.

We now specialize to undercomplete autoencoders, for which \eqref{eqn:bigQ_general_update} simplifies substantially. Let $\Sigma$ have eigenvalues $\lambda_{1} \geq \lambda_{2} \geq \cdots \geq \lambda_{d} \geq 0$ with a strict spectral gap $\lambda_{k} > \lambda_{k+1}$ separating the $k$-principal subspace from the remaining directions, and let $P \in \mathbb{R}^{k \times d}$ be an orthogonal projection to the principal subspace, with $PP^{\top}=I_{k}$. Then, the global loss minima form the manifold $(W,U) = (P^{\top}A,A^{-1} P)$ for $A \in \mathrm{GL}(k)$; in this sense linear autoencoders learn to implement PCA by projecting their inputs to the principal subspace \citep{baldi1989pca,kunin2019loss,pashakhanloo2025contribution}.

We next come to the key geometric fact. On the PCA solution manifold, the residual $r = (WU-I_{d}) x$ for an arbitrary datapoint $x$ lies entirely in the complement of the principal subspace, as $I_{d}-WU = I_{d} - PP^{\top} \equiv P_{\perp}$ projects to the complement. This forces $\nabla_{U}\hat{L}_{t}=0$ to be identically zero, while $\nabla_{W}\hat{L}_{t} = -P_{\perp} \hat{\Sigma}_{t} P^{\top} A_{t}^{-\top}$ depends on the product of in- and out-of-subspace fluctuations, and does not vanish (App.~\ref{app:pca_basics}). Thus, 
\begin{align}
    Q_{t+1} - Q_{t} = \eta^{2} (\nabla_{W}\hat{L}_{t})^{\top} (\nabla_{W}\hat{L}_{t}) \succeq 0
\end{align}
on the PCA solution manifold; $Q$ evolves in a preferred direction within the cone of PSD matrices. This is a directional bias towards decoder-heavy weights, as can be seen by considering the scalar charge $q = \tr(Q) = \Vert W \Vert_{F}^{2} - \Vert U \Vert_{F}^{2}$ associated with global rescalings $(W,U) \mapsto (a W, a^{-1} U)$ for $a > 0$ (Figure~\ref{fig2:workshop}A). No noise-balanced equilibrium, where the two terms in \eqref{eqn:bigQ_general_update} cancel on average \citep{ziyin2025noise}, is possible. In contrast, for an expansive autoencoder perfect reconstruction is possible, meaning that there is not a fundamental geometric reason for directed drift (Figure~\ref{fig2:workshop}A). We now analyze the scale dynamics.

\section{Effective scale dynamics}

We can derive a simple set of effective dynamics that allow us to predict the anisotropic growth in decoder scale. Under SGD with minibatch size $B$, changes to the balancing matrix $Q$ accumulate over $T_{\mathrm{slow}} \sim B/\eta^2$ steps, which is slow relative to the $T_{\mathrm{fast}} \sim 1/\eta$ steps over which fluctuations away from the solution manifold relax back towards it, as $T_{\mathrm{fast}}/T_{\mathrm{slow}} \sim \eta/B \ll 1$ when $\eta$ is small (see App.~\ref{app:model} for details). By leveraging this separation of timescales, we show in App.~\ref{app:model} that the dynamics of the matrix $S=A^{\top} A$ summarizing the scale degrees of freedom are given in the small-learning-rate limit $\eta\downarrow 0$ by the flow
\begin{align}
    \partial_{\tau} S + S^{-1} \partial_{\tau}S S^{-1} = S^{-1/2} C S^{-1/2},
\end{align}
where $C = \mathbb{E}[\Vert x_{\perp} \Vert^2 x_{\parallel}x_{\parallel}^{\top}]$ is defined in terms of $x_{\parallel}=Px$ and $x_{\perp} = (I_{d}-P^{\top}P)x$. If $S(\tau=0)$ and $C$ commute, then this matrix differential equation can be solved analytically. Working in a basis where $S = \diag(s_{1},\ldots,s_{k})$ and $C=\diag(c_1,\ldots,c_{k})$, we have the solution 
\begin{align}\label{eqn:model_scale_growth_main}
    s_{a}(\tau) = \sqrt{W[ s_{a}(0)^2 \exp(s_{a}(0)^2 + 2 c_{a} \tau) ]}, \quad a \in [k], 
\end{align}
where $W(\cdot)$ is the principal branch of the Lambert $W$-function \citep{dlmf}. Therefore, we expect anisotropic rates of growth along the different scale directions. In the Gaussian case, we have $c_{a} = \lambda_{a} \sum_{j=k+1}^{d}\lambda_{j}$, \textit{i.e.}, the rate of growth along the $a$-th direction is determined by the population variance along that direction multiplied by the total variance outside of the principal subspace. The fact that scale drift is driven by the coupling between within-subspace and out-of-subspace directions mirrors \citet{pashakhanloo2025contribution}'s finding that this same coupling governs the rate of drift along symmetry directions corresponding to rotations rather than re-scalings. This effective model produces excellent agreement with simulations at small but finite stepsize (Figure~\ref{fig2:workshop}C).

\section{Sharpness and the edge of stability}

Na\"ively, these effective dynamics suggest that the scales should grow indefinitely. However, this cannot be the case: indefinite scale growth would destabilize the underlying discrete SGD dynamics. In particular, the assumption that the dynamics relax stably back to the PCA manifold will fail if the eigenvalues $\mu$ of the population Hessian become too large---that is, if $\eta \mu$ exceeds some threshold $\kappa>0$, which for full-batch GD would be exactly equal to 2 \citep{cohen2021gradient}. For $S = \diag(s_1,\ldots,s_k)$ with ordered entries $s_1 > s_2 > \cdots > s_k$, we show in Appendix~\ref{app:model} that the Hessian of the population loss has dangerous modes $\mu_{11}^{\parallel} > \mu_{12}^{\parallel} > \cdots > \mu_{1k}^{\parallel}$ given by $\mu_{1a}^{\parallel} = \lambda_{1} ( s_{1}^{-1} + s_{a})$ for $a \in [k]$. Therefore, as the scales increase anisotropically, the curvature along each of these modes will increase until it hits the threshold $\eta \mu_{1a}^{\parallel} = \kappa$ at which the underlying discrete dynamics become unstable. Thus, we predict that all scales will eventually saturate at a common value given by $s_{a} = \kappa/\eta\lambda_1 + \mathcal{O}(\eta)$. We show that this prediction is consistent with simulations, with $\kappa \simeq 2$, in Figure~\ref{fig2:workshop}D-F. 

Consistent with the parameterization-dependence of the Euclidean Hessian \citep{dinh2017sharp}, not all notions of sharpness increase with scale. Most simply, considering global re-scalings with $S = s I_{k}$, the Hessian on the PCA manifold has trace $\tr(H) = \frac{d}{s} \sum_{j=1}^{k}\lambda_{j} + k s \sum_{j=1}^{d} \lambda_{j}$, which is minimized at $s_{\star}^2 = ({\frac{1}{k} \sum_{j=1}^{k}\lambda_{j}})/({\frac{1}{d}\sum_{j=1}^{d}\lambda_{j}}) \geq 1$ (App.~\ref{app:hessian}). More strikingly, \citet{andreyev2025stochasticstability} proposed a measure of ``batch sharpness'' that they suggest increases up to a threshold over SGD training; we show in Appendix~\ref{app:bs} that this measure is \textit{decreasing} in global scale on the PCA manifold. Moreover, we can prove that, for Gaussian data and a balanced initialization, their simplified ``modified batch sharpness'' measure is strictly decreasing under the effective dynamics, even as the population Hessian sharpens.

\section{Generalization to homogeneous nonlinear autoencoders}

Introducing nonlinear activation functions breaks the $\mathrm{GL}(k)$ symmetry of linear autoencoders. For homogeneous activation functions like ReLU, a neuron-wise scaling symmetry remains, with $A$ now being a diagonal matrix with positive diagonal entries rather than a general element of $\mathrm{GL}(k)$ (App.~\ref{app:nonlinear}). Our preliminary results suggest that undercomplete ReLU autoencoders show a similar directed drift towards decoder-heavy weights (Fig.~\ref{fig:relu-scale-imbalance}). We leave an in-depth investigation of the geometric basis for this bias to future work.

\section{Discussion}

We have shown that undercomplete linear autoencoders trained with SGD display directed scale drift towards decoder-heavy weights. The endpoint of this drift is selected not by an equilibrium statistical mechanism like balancing of gradient noise \citep{ziyin2025noise}, but instead dynamically by the fact that scale drift increases curvature up to the point where the solution would become unstable. 

From a broader optimization perspective, this resembles the general phenomenon of \textit{progressive sharpening} in full-batch gradient descent, which posits that the top eigenvalues of the Hessian increase sequentially over training up to the stability threshold, after which they remain near the edge of stability \citep{cohen2021gradient,cohen2025understanding,damian2023self}. How to precisely define the edge of stability in the stochastic setting remains an active area of research \citep{andreyev2025stochasticstability,emmanouilidis2026edge}. What is notable in our setting is the fact that the edge of stability is reached purely through a slow effective drift along a manifold of functionally-equivalent solutions, after the PCA solution manifold is reached. Therefore, progressive sharpening need not reflect continued learning of the function. Whether and how this phenomenology transfers to other optimization settings is an interesting direction for future work.

\acks{F.P. was supported by the Harvard Center for Brain Science (CBS)-NTT Fellowship Program on the Physics of Intelligence. J.A.Z.-V. and this research were supported by the Office of the Director of the NIH under Award Number DP5OD037354, and additionally by a Junior Fellowship from the Harvard Society of Fellows.}

\section*{Declaration of LLM Usage}
The development of this research did not involve LLMs in its core aspects. However, we used LLMs (OpenAI's Codex) to generate codes for the simulations results presented here. We subsequently checked the generated code and plots.

\clearpage
\bibliography{refs}

\clearpage 

\appendix
\numberwithin{equation}{section}
\setlength{\abovedisplayskip}{0.1em}
\setlength{\belowdisplayskip}{0.1em}

\renewcommand{\thefigure}{S\arabic{figure}}
\setcounter{figure}{0}

\section{Linear autoencoder basics}\label{app:pca_basics}

Here, we give a detailed discussion of the basic setup for linear autoencoders. Recall that the basic model architecture is
\begin{align}
    f(x) = W Ux,
\end{align}
where $U \in \mathbb{R}^{k \times d}$ and $W \in \mathbb{R}^{d \times k}$. From the definition of the model, we can see that there is a $\mathrm{GL}(k)$ symmetry: the function output is invariant under transformations of the weights of the form
\begin{align}
    (W,U) \mapsto (W A, A^{-1} U), \quad A \in \mathrm{GL}(k).
\end{align}
Associated with this symmetry is the Noether charge \citep{kunin2021neural,tanaka2021noether}
\begin{align}
    Q = W^{\top}W - UU^{\top}.
\end{align}

\subsection{Population loss and gradient flow}

For a set of data with population covariance $\mathbb{E}[xx^{\top}] = \Sigma$, the population reconstruction loss is 
\begin{align}
    L &= \frac{1}{2} \mathbb{E}_{x}[\Vert f(x)-x \Vert^2] 
    = \frac{1}{2} \mathbb{E}_{x}\tr[(WU-I_{d}) xx^{\top} (WU-I_{d})^{\top}] 
    \\&= \frac{1}{2} \tr[(WU-I_{d}) \Sigma (WU-I_{d})^{\top}] .
\end{align}
Under population gradient flow, the parameters then evolve as
\begin{align}
    \partial_{t} W &= - \nabla_{W} L = -(WU-I_{d}) \Sigma U^{\top}
    \\
    \partial_{t} U &= - \nabla_{U} L = -W^{\top} (WU-I_{d}) \Sigma.
\end{align}
The charge $Q = W^{\top}W - UU^{\top}$ is conserved under these dynamics: 
\begin{align}
    \partial_{t} Q &= (\partial_{t} W)^{\top} W + W^{\top} \partial_{t} W - (\partial_{t} U) U^{\top} - U (\partial_{t} U)^{\top} = 0,
\end{align}
as we can see that $W^{\top} \partial_{t} W = (\partial_{t} U) U^{\top}$. 

\subsection{Minibatch stochastic gradient descent}

Now consider stochastic gradient descent with minibatch size $B$. Let the matrix of data in the minibatch at time $t$ be $X_{t} \in \mathbb{R}^{d \times B}$, and 
\begin{align}
    \hat{\Sigma}_{t} = \frac{1}{B} X_{t} X_{t}^{\top}
\end{align}
be the empirical covariance of the minibatch. Then, the minibatch loss is 
\begin{align}
    \hat{L}_{t} = \frac{1}{2} \tr[(WU-I_{d}) \hat{\Sigma}_{t} (WU-I_{d})^{\top}], 
\end{align}
so the minibatch gradients are
\begin{align}
    \nabla_{W}\hat{L}_{t} &= (W_{t} U_{t} - I_{d}) \hat{\Sigma}_{t} U_{t}^{\top} \\ 
    \nabla_{U}\hat{L}_{t} &= W_{t}^{\top} (W_{t} U_{t} - I_{d}) \hat{\Sigma}_{t} .
\end{align}
and the update to the parameters is
\begin{align}
    W_{t+1} &= W_{t} - \eta \nabla_{W}\hat{L}_{t} = W_{t} - \eta (W_{t} U_{t} - I_{d}) \hat{\Sigma}_{t} U_{t}^{\top} \\ 
    U_{t+1} &= U_{t} - \eta \nabla_{U} \hat{L}_{t} = U_{t} - \eta W_{t}^{\top} (W_{t} U_{t} - I_{d}) \hat{\Sigma}_{t} .
\end{align}
Using the fact that $W_{t}^{\top} \nabla_{W}\hat{L}_{t} = (\nabla_{U}\hat{L}_{t}) U^{\top}$, we find that the corresponding update to the balancing matrix $Q_{t} = W_{t}^{\top} W_{t} - U_{t} U_{t}^{\top}$ is
\begin{align}
    Q_{t+1} - Q_{t} &= \eta^{2} \bigg( (\nabla_{W}\hat{L}_{t})^{\top} (\nabla_{W}\hat{L}_{t}) - (\nabla_{U}\hat{L}_{t}) (\nabla_{U}\hat{L}_{t})^{\top} \bigg). 
\end{align}
Substituting in the definition of $\nabla_{W}\hat{L}_{t}$ and $\nabla_{U}\hat{L}_{t}$ in terms of $\hat{\Sigma}_{t}$, and defining the residual 
\begin{align}
    R_{t} = (W_{t}U_{t}-I_{d}) X_{t},
\end{align}
we can re-write this as 
\begin{align}
    Q_{t+1} = Q_{t} + \frac{\eta^{2}}{B^2} (U_{t} X_{t} R_{t}^{\top} R_{t} X_{t}^{\top} U_{t}^{\top} - W_{t}^{\top} R_{t} X_{t} X_{t}^{\top} R_{t}^{\top} W_{t} ).
\end{align}

\subsection{PCA solutions}

Our interest is in PCA-type solutions \citep{baldi1989pca,kunin2019loss,pashakhanloo2025contribution}
\begin{align}
    (W,U) = (P^{\top} A, A^{-1} P),
\end{align}
where $P \in \mathbb{R}^{k \times d}$ is an orthogonal projection $PP^{\top} = I_{k}$ and $A \in \mathrm{GL}(k)$ includes the symmetry degrees of freedom. For such a solution, 
\begin{align}
    I_{d} - W U = I_{d} - P^{\top} P = P_{\perp}
\end{align}
is a projection onto the orthogonal complement of the span of $P^{\top}$. As a result, 
\begin{align}
    W^{\top} (W U - I_{d} ) = A^{\top} P (P^{\top} P - I_{d}) = A^{\top} (P-P) = 0
\end{align}
as $PP^{\top} = I_{k}$, meaning that 
\begin{align}
    \nabla_{U} \hat{L}_{t} = 0
\end{align}
on the PCA solution manifold. Importantly, this is a statement that the gradient is \emph{identically} zero on the solution manifold. In contrast, 
\begin{align}
    \nabla_{W} \hat{L}_{t} = (W_{t}U_{t}-I_{d}) \hat{\Sigma}_{t} U_{t}^{\top} = (P^{\top} P-I_{d}) \hat{\Sigma}_{t} P^{\top} A_{t}^{-\top}
\end{align}
is not identically zero. Indeed, if we introduce the decomposition 
\begin{align}
    X_{t} = P^{\top} X_{t,\parallel} + X_{t,\perp}
\end{align}
such that
\begin{align}
    P X_{t} = X_{t,\parallel} \quad \textrm{and} \quad P_{\perp} X_{t} = X_{t,\perp},
\end{align}
we find that
\begin{align}
    \nabla_{W} \hat{L}_{t} = - \frac{1}{B} X_{t,\perp} X_{t,\parallel}^{\top} A_{t}^{-\top}.
\end{align}
Together, this leads to 
\begin{align}
    Q_{t+1}-Q_{t} 
    &= \eta^{2} (\nabla_{W}\hat{L}_{t})^{\top} (\nabla_{W}\hat{L}_{t})
    = \frac{\eta^2}{B^2} A_{t}^{-1} X_{t,\parallel} X_{t,\perp}^{\top}X_{t,\perp} X_{t,\parallel}^{\top} A_{t}^{-\top}
    \succeq 0
\end{align}
on the PCA solution manifold.

\section{Spectrum of the linear autoencoder Hessian}\label{app:hessian}

We now derive the Hessian of the linear autoencoder reconstruction loss. This computation largely parallels previous Hessian computations by \citet{baldi1989pca,pashakhanloo2023drift,pashakhanloo2025contribution}. 

Our general assumption on the data is that 
\begin{align}
    \mathbb{E}[x] = 0 \quad \mathrm{and} \quad \mathbb{E}[xx^{\top}] = \Sigma,
\end{align}
where $\Sigma$ has an orthogonal eigendecomposition
\begin{align}
    \Sigma = \sum_{i=1}^{d} \lambda_{i} v_{i} v_{i}^{\top},
\end{align}
with $\lambda_{1} \geq \lambda_{2} \geq \cdots \geq \lambda_{d} \geq 0$ and $v_{i}^{\top} v_{j} = \delta_{ij}$. Without loss of generality, we may assume that the eigenvectors $v_{i}$ are the standard Euclidean basis vectors; we will impose this assumption when convenient. We will also often assume that the eigenvalues satisfy spectral gap conditions of the form $\lambda_{k} > \lambda_{k+1}$ for some fixed $k$; this is required for the $k$-principal subspace to be well-defined.

\subsection{Warm-up: Computing the top PC}

As a warm-up, consider the case $k=1$, in which we want to compute the top principal component. Then, the loss is 
\begin{align}
    L &= \frac{1}{2} \mathbb{E}_{x} \Vert w u^{\top} x - x\Vert^{2} 
    = \frac{1}{2} \tr[(wu^{\top}-I_{d}) \Sigma (uw^{\top}-I_{d})] \\& = \frac{1}{2} \Vert w \Vert^2 (u^{\top} \Sigma u) - u^{\top} \Sigma w + \frac{1}{2} \tr(\Sigma),
\end{align}
from which we can easily see the Hessian is
\begin{align}
    H = \begin{pmatrix}
        (u^{\top} \Sigma u) I_{d} & 2 w u^{\top} \Sigma - \Sigma 
        \\
        2 \Sigma u w^{\top} - \Sigma & \Vert w \Vert^2 \Sigma
    \end{pmatrix}.
\end{align}
 Consider a general PCA-type solution for some fixed $i \in [d]$ and a scale factor $s>0$:
\begin{align}
    w = s^{1/2} v_{i}, \quad u = s^{-1/2} v_{i} ,
\end{align}
which projects the data onto the $i$-th principal direction. Then, the Hessian becomes
\begin{align}
    H = 
    \begin{pmatrix}
        \lambda_{i} s^{-1} I_{d} & \lambda_{i} v_{i} v_{i}^{\top} - \sum_{j \neq i} \lambda_{j} v_{j} v_{j}^{\top}
        \\
        \lambda_{i} v_{i} v_{i}^{\top} - \sum_{j \neq i} \lambda_{j} v_{j} v_{j}^{\top} & s \sum_{j} \lambda_{j} v_{j} v_{j}^{\top}
    \end{pmatrix}. 
\end{align}
As a result, if we expand a general perturbation as 
\begin{align}
    \begin{pmatrix}
        \delta w \\ \delta u 
    \end{pmatrix}
    = \begin{pmatrix}
        \sum_{j=1}^{d} \alpha_{j} v_{j} \\ \sum_{j=1}^{d} \beta_{j} v_{j} 
    \end{pmatrix},
\end{align}
we have the expansion
\begin{align}
    \begin{pmatrix}
        \delta w \\ \delta u 
    \end{pmatrix}^{\top} H \begin{pmatrix}
        \delta w \\ \delta u 
    \end{pmatrix} 
    = \sum_{j=1}^{d} \begin{pmatrix} \alpha_{j} \\ \beta_{j} \end{pmatrix}^{\top} H_{j} \begin{pmatrix} \alpha_{j} \\ \beta_{j} \end{pmatrix},
\end{align}
where
\begin{align}
    H_{i} = \lambda_{i} \begin{pmatrix}
         s^{-1} & 1
        \\ 
        1 & s
    \end{pmatrix}
\end{align}
and, for $j \neq i$, 
\begin{align}
    H_{j} &= 
    \begin{pmatrix}
        \lambda_{i} s^{-1} & - \lambda_{j} 
        \\ 
        - \lambda_{j} & s \lambda_{j}
    \end{pmatrix}.
\end{align}
This means that we may equivalently write
\begin{align}
    \begin{pmatrix}
        H_{1} & &  \\
         & H_{2} &  \\ 
         &  & \ddots 
    \end{pmatrix} 
    \begin{pmatrix} \alpha_{1}\\ \beta_{1} \\ \alpha_{2} \\ \beta_{2} \\ \vdots \end{pmatrix}
\end{align}
in these coordinates. This means that we can diagonalize the full Hessian by diagonalizing the $2 \times 2$ blocks. 

The block $H_{i}$, which corresponds to fluctuations in the principal subspace, has eigenvalue $\mu_{i,-}=0$, corresponding to the scale symmetry and the eigenvector
\begin{align}
    \frac{1}{\sqrt{s+s^{-1}}}
    \begin{pmatrix}
        s^{1/2} \\ -s^{-1/2}
    \end{pmatrix}
\end{align}
and 
\begin{align}
    \mu_{i,+} = \lambda_{i} (s + s^{-1}),
\end{align}
corresponding to the eigenvector
\begin{align}
    \frac{1}{\sqrt{s+s^{-1}}}
    \begin{pmatrix}
        s^{1/2} \\ s^{-1/2}
    \end{pmatrix}.
\end{align}
For $j \neq i$, we have that
\begin{align}
    \det H_{j} = \lambda_{j} (\lambda_{i} - \lambda_{j}),
\end{align}
and eigenvalues
\begin{align}
    \mu_{j,\pm} = \frac{\lambda_{i} s^{-1} + \lambda_{j} s  \pm \sqrt{(\lambda_{i} s^{-1} + \lambda_{j} s )^2 - 4 \lambda_{j}(\lambda_{i}-\lambda_{j})}}{2}
\end{align}
corresponding to eigenvectors 
\begin{align}
    \frac{1}{\sqrt{\lambda_{j}^2 + (\mu_{j,\pm} - \lambda_{i} s^{-1})^2}} \begin{pmatrix} - \lambda_{j} \\ \mu_{j,\pm} - \lambda_{i} s^{-1} \end{pmatrix}.
\end{align}

This allows us to see several things. First, the Hessian is positive-semidefinite only if $\lambda_{i} \geq \lambda_{j}$ for all $j \neq i$, and positive-definite on the orthogonal complement of the direction corresponding to the scaling symmetry only if $\lambda_{i} > \lambda_{j} > 0$. This is what we expect from the PCA solution. 

We now consider the effect of rescaling. First, it is clear that $\mu_{i,+} = \lambda_{i} (s + s^{-1})$ is always increased by breaking the scale balance, but it behaves identically irrespective of whether the solution is decoder-heavy or encoder-heavy. 

Now, for $j \neq i$, we have 
\begin{align}
    \mu_{j,\pm} = \frac{\lambda_{i}}{2} \left[ s^{-1} + \tilde{\lambda}_{j} s \pm \sqrt{\left(s^{-1} + \tilde{\lambda}_{j} s \right)^2 - 4 \tilde{\lambda}_{j} (1-\tilde{\lambda}_{j})} \right] 
\end{align}
after factoring out $\lambda_i>0$ and defining
\begin{align}
    \tilde{\lambda}_{j} = \frac{\lambda_j}{\lambda_i} .
\end{align}
Assuming $\tilde{\lambda}_j>0$, the function
\begin{align}
    t(s) = \frac{1}{\lambda_i} \tr(H_{j}) = s^{-1} + \tilde{\lambda}_{j} s
\end{align}
has a unique minimum over $s>0$ at
\begin{align}
    s_{\star} = \tilde{\lambda}_{j}^{-1/2},
\end{align}
where it takes the value $t(s_{\star}) = 2 \tilde{\lambda}_{j}^{1/2}$. From this, we find that $\mu_{j,+}$ decreases in $s$ on $(0,\tilde{\lambda}_{j}^{-1/2})$, has a global minimum at $s_{\star} = \tilde{\lambda}_{j}^{-1/2}$ where it takes value
\begin{align}
    \mu_{j,+} \bigg|_{s_{\star} = \tilde{\lambda}_{j}^{-1/2}} = \lambda_{i} (\tilde{\lambda}_{j}^{1/2} + \tilde{\lambda}_{j}),
\end{align}
and increases on $(\tilde{\lambda}_{j}^{-1/2},\infty)$. Analogously, $\mu_{j,-}$ increases on $(0,\tilde{\lambda}_{j}^{-1/2})$, has a global maximum at $s_{\star} = \tilde{\lambda}_{j}^{-1/2}$ where it takes value
\begin{align}
    \mu_{j,-} \bigg|_{s_{\star} = \tilde{\lambda}_{j}^{-1/2}} = \lambda_{i} (\tilde{\lambda}_{j}^{1/2} - \tilde{\lambda}_{j}),
\end{align}
and decreases on $(\tilde{\lambda}_{j}^{-1/2},\infty)$. This shows that changing the scale has, in general, a rather non-obvious effect on the spectrum of the Hessian, even in this simple case.

We now prove several statements about the ordering of the eigenvalues. To do so, we first derive several preliminary results. Let
\begin{align}
    \mu_{\pm}(\lambda) = \frac{1}{2} [s^{-1} + \lambda s \pm \sqrt{(s^{-1}+\lambda s)^2 - 4 \lambda (1-\lambda)}]
\end{align}
be functions defining the normalized eigenvalues, with $s>0$ fixed and $\lambda \in [0,1]$. By inspection, we have $\mu_{\pm}(\lambda) \geq 0$ for all $\lambda \in [0,1]$. 

First consider $\mu_{+}(\lambda)$. At the boundaries, we have
\begin{align}
    \mu_{+}(\lambda=0) = s^{-1} \quad \textrm{and} \quad \mu_{+}(\lambda=1) = s + s^{-1}. 
\end{align}
Moreover, differentiating in $\lambda$, we have
\begin{align}
    \mu_{+}'(\lambda) = \frac{1}{2} \left(s +  \frac{(s^{2}+4)\lambda-1}{\sqrt{(s^{-1}+\lambda s)^2 - 4 \lambda (1-\lambda)}} \right),
\end{align}
which is positive on the interval $\lambda \in (0,1)$ for any $s$. To show this, consider two cases. First, if the numerator of the fraction inside the square brackets is non-negative, than the result is immediate. If the numerator is negative, then we must have
\begin{align}
    0 < s^{2}[(s^{-1}+\lambda s)^2 - 4 \lambda (1-\lambda)] - [(s^{2}+4)\lambda-1]^2 
    = 4 \lambda [2 - (s^{2}+4) \lambda]. 
\end{align}
But, for the numerator to be negative, we must have $\lambda < 1/(s^{2}+4)$, meaning that $2 - (s^{2}+4) \lambda > 1$, and thus that the desired inequality is in fact satisfied. 

Now consider $\mu_{-}(\lambda)$. At the boundaries, we see that
\begin{align}
    \mu_{-}(\lambda=0) = \mu_{-}(\lambda=1) = 0.
\end{align}
By differentiating, we find that $\mu_{-}(\lambda)$ has a unique maximum in $\lambda$ at
\begin{align}
    \lambda_{\star} = \frac{2}{s^{2}+4},
\end{align}
where it takes value
\begin{align}
    \mu_{-}(\lambda_{\star}) = \frac{s}{s^{2}+4}. 
\end{align}
Note that for any $s>0$, 
\begin{align}
    \frac{s}{s^{2}+4} < s^{-1},
\end{align}
as upon multiplying through by $s (s^{2}+4)$ this is equivalent to the inequality $s^{2} < s^{2}+4$. 

Therefore, assuming that $i=1$ is the unique maximal eigenvalue, we conclude three things about the functions $\mu_{\pm}(\lambda)$: $\mu_{+}(\lambda)$ is monotonically increasing, $\mu_{-}(\lambda)$ is non-monotonic, and we have the strict uniform-bound separation
\begin{align}
    0 \leq \mu_{-}(\lambda) \leq \frac{s}{s^{2}+4} < s^{-1} \leq \mu_{+}(\lambda') \leq s+s^{-1}
\end{align}
for any pair $\lambda,\lambda' \in [0,1]$. 

With this lemma in hand, we find that for any pair $(j,k) \in [d]^2$ such that $\lambda_{j} > \lambda_{k}$, the corresponding upper-branch eigenvalues of the Hessian satisfy 
\begin{align}
    \mu_{j,+} > \mu_{k,+}. 
\end{align}
Moreover, we have a strict separation between the two branches: for any pair $(j,k)$ we have that
\begin{align}
    \frac{1}{\lambda_1} \mu_{j,-} \leq \frac{s}{s^{2}+4} < s^{-1} \leq \frac{1}{\lambda_1} \mu_{k,+}.
\end{align}
However, the lower-branch eigenvalues $\mu_{j,-}$ are not ordered simply by the corresponding eigenvalues $\lambda_{j}$ of $\Sigma$. We remark that this result guarantees strong separation of bands when $s<1$, but allows the gap to in principle become small when $s \gg 1$. 

As a result, assuming that $i=1$ is the index of the maximum eigenvalue, the leading eigenvalue of the Hessian is always $\mu_{i,+} = \lambda_{i} (s+s^{-1})$. Thus, the leading eigenvalue is minimized by the balanced solution, and in this sense balancing minimizes sharpness. However, it does not favor imbalance in a particular direction, as it is invariant under $s \mapsto 1/s$. Asymmetric effects on the Hessian therefore must come from lower eigenvalues, which are divided into two non-interlacing bands. 

One way to capture the asymmetry is through the trace of the Hessian:
\begin{align}
    \tr(H) &= \tr(H_{i}) + \sum_{j \neq i} \tr(H_j)
    \\
    &= \lambda_{i} (s + s^{-1}) + \sum_{j \neq i} ( \lambda_{j} s + \lambda_{i} s^{-1} )
    \\
    &= \tr(\Sigma) s + d \lambda_{i} s^{-1} .
\end{align}
We can then see that the trace is invariant under $s \mapsto 1/s$ only if $\lambda_{i} = \lambda_{j}$ for all $j$, \textit{i.e.}, if all the eigenvalues of $\Sigma$ are identical. Otherwise, we have the strict inequality
\begin{align}
    d \lambda_{i} > \tr(\Sigma) ,
\end{align}
which means that the trace of the Hessian increases more as one moves towards a encoder-heavy solution ($s<1$) as compared to a decoder-heavy solution ($s>1$). We note that $\tr(H)$ is minimized by taking
\begin{align}
    s_{\star} = \left(\frac{\tr(\Sigma)}{d\lambda_{i}}\right)^{-1/2} \geq 1;
\end{align}
it is decreasing on $(0,s_{\star})$ and increasing on $(s_{\star},\infty)$. This means that in the sense of $\tr(H)$, decoder-heavy solutions can be flatter than encoder-heavy solutions or even balanced solutions.

\subsection{Hessian spectrum for a general undercomplete linear autoencoder}

We now consider a general undercomplete linear autoencoder with $k > 1$. We will find that the Hessian has a similar structure to the $k=1$ case---being divided into sectors corresponding to within-principal-subspace and out-of-subspace fluctuations. 

Recall that the loss, after averaging over some set of examples with $\mathbb{E}[x]=0$ and $\mathbb{E}[x x^{\top}] = \Sigma$, is 
\begin{align}
    L(W,U) 
    = \mathbb{E}_{x}\frac{1}{2} \Vert W U x-x\Vert^{2}
    = \frac{1}{2} \tr[(WU-I_d) \Sigma (U^{\top}W^{\top}-I_d)] . 
\end{align}
Consider a generic point $(W,U)$, and perturb this point to $(W+\delta W, U+\delta U)$. Then, 
\begin{align}
    &L(W+\delta W, U+\delta U)
    \nonumber\\&= \frac{1}{2} \tr[(WU + W \delta U + \delta W U + \delta W \delta U -I_d) \Sigma \nonumber\\&\qquad\qquad \times (U^{\top}W^{\top} + \delta U^{\top} W^{\top} + U^{\top} \delta W^{\top} + \delta U^{\top} \delta W^{\top}-I_d)]
    \\
    &= L(W,U) 
    + \tr[( W \delta U + \delta W U ) \Sigma (WU - I_{d} )^{\top}]
    \nonumber\\&\quad + \frac{1}{2} \tr[( W \delta U + \delta W U ) \Sigma ( \delta U^{\top} W^{\top} + U^{\top} \delta W^{\top}  )]   + \tr[( \delta W \delta U ) \Sigma (WU - I_{d} )^{\top}]
    \nonumber\\&\quad + \textrm{cubic and quartic terms}.
\end{align}
As a result, viewing the Hessian as a quadratic form acting on the perturbation $(\delta W,\delta U)$, we have
\begin{align}
    H(\delta W,\delta U)
    &= \tr[( W \delta U + \delta W U ) \Sigma ( W \delta U + \delta W U )^{\top}]   + 2 \tr[( \delta W \delta U ) \Sigma (WU - I_{d} )^{\top}]. 
\end{align}

Now, we specialize to a rank-$k$ undercomplete PCA-type solution, with
\begin{align}
    W &= P^{\top} A,
    \\
    U &= A^{-1} P,
\end{align}
where $A$ is a general invertible matrix and $P \in \mathbb{R}^{k \times d}$ is a projection matrix satisfying $P P^{\top} = I_{k}$. Specifically, we assume that that $P$ projects onto the eigenspace spanned by the top $k$ eigenvectors, assuming that it is unique (\textit{i.e.}, that $\lambda_{k} > \lambda_{k+1}$). 

Choose coordinates for the input space such that $\Sigma$ is diagonal, and then for the latent space such that
\begin{align}
    P = \begin{pmatrix}
        I_{k} & 0
    \end{pmatrix}. 
\end{align}
In these coordinates, decompose the perturbations as 
\begin{align}
    \delta W = \begin{pmatrix} \delta W_{\parallel} \\ \delta W_{\perp} \end{pmatrix} \quad \mathrm{and} \quad 
    \delta U = \begin{pmatrix} \delta U_{\parallel} & \delta U_{\perp} \end{pmatrix},
\end{align}
where 
\begin{align}
    \delta W_{\parallel},\delta U_{\parallel} \in \mathbb{R}^{k \times k}
    \quad \mathrm{and} \quad 
    \delta W_{\perp}^{\top}, \delta U_{\perp} \in \mathbb{R}^{k \times (d-k)}.
\end{align}
With this choice of coordinates, 
\begin{align}
    W \delta U + \delta W U 
    &= \begin{pmatrix} A \\ 0 \end{pmatrix} \begin{pmatrix} \delta U_{\parallel} & \delta U_{\perp} \end{pmatrix} + \begin{pmatrix} \delta W_{\parallel} \\ \delta W_{\perp} \end{pmatrix} \begin{pmatrix} A^{-1} & 0 \end{pmatrix}
    \\
    &= \begin{pmatrix}
        \delta W_{\parallel} A^{-1} + A \delta U_{\parallel} & A \delta U_{\perp} \\ \delta W_{\perp} A^{-1} & 0
    \end{pmatrix}. 
\end{align}
Then, letting $\Lambda_{+} = \diag(\lambda_{1},\ldots,\lambda_{k})$ and $\Lambda_{-} = \diag(\lambda_{k+1},\ldots,\lambda_{d})$, we have
\begin{align}
    &\tr[( W \delta U + \delta W U ) \Sigma ( W \delta U + \delta W U )^{\top}]
    \\
    &= \tr\left[  \begin{pmatrix}
        \delta W_{\parallel} A^{-1} + A \delta U_{\parallel} & A \delta U_{\perp} \\ \delta W_{\perp} A^{-1} & 0
    \end{pmatrix} \begin{pmatrix} \Lambda_{+} & 0 \\ 0 & \Lambda_{-} \end{pmatrix}  \begin{pmatrix}
        (\delta W_{\parallel} A^{-1} + A \delta U_{\parallel})^{\top} & (\delta W_{\perp} A^{-1})^{\top} \\ (A \delta U_{\perp})^{\top} & 0
    \end{pmatrix} \right]
    \\
    &= \tr\left[  \begin{pmatrix}
        \delta W_{\parallel} A^{-1} + A \delta U_{\parallel} & A \delta U_{\perp} \\ \delta W_{\perp} A^{-1} & 0
    \end{pmatrix} \begin{pmatrix} \Lambda_{+} (\delta W_{\parallel} A^{-1} + A \delta U_{\parallel})^{\top} & \Lambda_{+} (\delta W_{\perp} A^{-1})^{\top} \\ \Lambda_{-} (A \delta U_{\perp})^{\top} & 0 \end{pmatrix} \right]
    \\
    &= \tr[(\delta W_{\parallel} A^{-1} + A \delta U_{\parallel}) \Lambda_{+} (\delta W_{\parallel} A^{-1} + A \delta U_{\parallel})^{\top}]
    \nonumber\\&\quad + \tr[A \delta U_{\perp} \Lambda_{-} (A \delta U_{\perp})^{\top}]
    \nonumber\\&\quad + \tr[\delta W_{\perp} A^{-1} \Lambda_{+} A^{-\top} \delta W_{\perp}^{\top}]
\end{align}
and, as 
\begin{align}
    \Sigma (WU-I_d)^{\top} = \begin{pmatrix}
        0 & 0 \\ 0 & -\Lambda_{-} 
    \end{pmatrix},
\end{align}
\begin{align}
    \tr[( \delta W \delta U ) \Sigma (WU - I_{d} )^{\top}]
    &= -\tr[\delta W_{\perp} \delta U_{\perp} \Lambda_{-}] . 
\end{align}

Therefore, the Hessian decomposes into two independent quadratic forms 
\begin{align}
    H(\delta W,\delta U) &= H_{\parallel}(\delta W_{\parallel}, \delta U_{\parallel}) + H_{\perp}(\delta W_{\perp},\delta U_{\perp}),
\end{align}
with 
\begin{align}
    H_{\parallel}(\delta W_{\parallel}, \delta U_{\parallel}) &= \tr[(\delta W_{\parallel} A^{-1} + A \delta U_{\parallel}) \Lambda_{+} (\delta W_{\parallel} A^{-1} + A \delta U_{\parallel})^{\top}]
\end{align}
determined by within-subspace fluctuations, and
\begin{align}
    H_{\perp}(\delta W_{\perp},\delta U_{\perp}) = \tr[A \delta U_{\perp} \Lambda_{-} \delta U_{\perp}^{\top} A^{\top}]  + \tr[\delta W_{\perp} A^{-1} \Lambda_{+} A^{-\top} \delta W_{\perp}^{\top}] - 2\tr[\delta W_{\perp} \delta U_{\perp} \Lambda_{-}]
\end{align}
determined by out-of-subspace fluctuations. We can therefore diagonalize each separately. 

First, consider the out-of-subspace fluctuations. Using the vectorization convention 
\begin{align}
    \tr(A^{\top} B) = \vectorize(A)^{\top} \vectorize(B), \quad \vectorize(ABC) = (C^{\top} \otimes A) \vectorize(B) ,
\end{align}
we can write the three relevant terms as 
\begin{align}
     \tr[A \delta U_{\perp} \Lambda_{-} \delta U_{\perp}^{\top} A^{\top}]
     &= \vectorize(\delta U_{\perp})^{\top} (\Lambda_{-} \otimes A^{\top} A) \vectorize(\delta U_{\perp})
     \\
     \tr[\delta W_{\perp} A^{-1} \Lambda_{+} A^{-\top} \delta W_{\perp}^{\top}]
     &= \vectorize(\delta W_{\perp}^{\top})^{\top} (I_{d-k} \otimes A^{-1} \Lambda_{+} A^{-\top} ) \vectorize(\delta W_{\perp}^{\top})
     \\
     \tr[\delta W_{\perp} \delta U_{\perp} \Lambda_{-}] 
     &= \vectorize(\delta W_{\perp}^{\top})^{\top} (\Lambda_{-} \otimes I_{k}) \vectorize(\delta U_{\perp}), 
\end{align}
hence
\begin{align}
    H_{\perp}(\delta W_{\perp},\delta U_{\perp}) =
    \begin{pmatrix}
        \vectorize(\delta W_{\perp}^{\top}) \\ \vectorize(\delta U_{\perp}) 
    \end{pmatrix}^{\top}
    \begin{pmatrix}
        I_{d-k} \otimes A^{-1} \Lambda_{+} A^{-\top} & - \Lambda_{-} \otimes I_{k}
        \\
        - \Lambda_{-} \otimes I_{k} & \Lambda_{-} \otimes A^{\top} A
    \end{pmatrix} \begin{pmatrix}
        \vectorize(\delta W_{\perp}^{\top}) \\ \vectorize(\delta U_{\perp}) 
    \end{pmatrix}.
\end{align}
We can then see that there is a further splitting into eigenproblems for the $2k \times 2k$ block matrices
\begin{align}
    \begin{pmatrix}
        A^{-1} \Lambda_{+} A^{-\top} & - \lambda_{j} I_{k} \\ 
        - \lambda_{j} I_{k} & \lambda_{j} A^{\top} A
    \end{pmatrix}, 
\end{align}
for each $j \in \{k+1,\ldots,d\}$. This generalizes our result from the $k=1$ case before. 

We now at last come to $H_{\parallel}$. Expanding out all of the terms, 
\begin{align}
    H_{\parallel}
    &= \tr[(\delta W_{\parallel} A^{-1} + A \delta U_{\parallel}) \Lambda_{+} (\delta W_{\parallel} A^{-1} + A \delta U_{\parallel})^{\top}]
    \\
    &= \tr[(\delta W_{\parallel} A^{-1} + A \delta U_{\parallel}) \Lambda_{+} (A^{-\top} \delta W_{\parallel}^{\top} + \delta U_{\parallel}^{\top} A^{\top})]
    \\
    &= \tr\bigg[ \delta W_{\parallel} A^{-1} \Lambda_{+} A^{-\top} \delta W_{\parallel}^{\top} + \delta W_{\parallel} A^{-1} \Lambda_{+} \delta U_{\parallel}^{\top} A^{\top} + A \delta U_{\parallel} \Lambda_{+} A^{-\top} \delta W_{\parallel}^{\top} + A \delta U_{\parallel} \Lambda_{+} \delta U_{\parallel}^{\top} A^{\top} \bigg]
    \\
    &= \vectorize(\delta W_{\parallel})^{\top} (A^{-1} \Lambda_{+} A^{-\top} \otimes I_{k}) \vectorize(\delta W_{\parallel}) + 2 \vectorize(\delta W_{\parallel})^{\top} (A^{-1} \Lambda_{+} \otimes A) \vectorize(\delta U_{\parallel}) \nonumber\\&\quad + \vectorize(\delta U_{\parallel})^{\top} (\Lambda_{+} \otimes A^{\top} A) \vectorize(\delta U_{\parallel}),
\end{align}
which means that 
\begin{align}
    H_{\parallel}(\delta W_{\parallel}, \delta U_{\parallel})
    &= \begin{pmatrix}
        \vectorize(\delta W_{\parallel}) \\ \vectorize(\delta U_{\parallel})
    \end{pmatrix}^{\top}
    \begin{pmatrix}
        A^{-1} \Lambda_{+} A^{-\top} \otimes I_{k} & A^{-1} \Lambda_{+} \otimes A \\ \Lambda_{+} A^{-\top} \otimes A^{\top} & \Lambda_{+} \otimes A^{\top} A
    \end{pmatrix}
    \begin{pmatrix}
        \vectorize(\delta W_{\parallel}) \\ \vectorize(\delta U_{\parallel})
    \end{pmatrix},
\end{align}
and thus the remaining eigenvalues are determined by the spectrum of the $2k^2 \times 2k^2$ matrix
\begin{align}
    \begin{pmatrix}
        A^{-1} \Lambda_{+} A^{-\top} \otimes I_{k} & A^{-1} \Lambda_{+} \otimes A \\ \Lambda_{+} A^{-\top} \otimes A^{\top} & \Lambda_{+} \otimes A^{\top} A
    \end{pmatrix}.
\end{align}
We note that we can factorize this matrix as $J^{\top} J$, where 
\begin{align}
    J = \begin{pmatrix} 
            \Lambda_{+}^{1/2} A^{-\top} \otimes I_{k} & \Lambda_{+}^{1/2} \otimes A
        \end{pmatrix} \in \mathbb{R}^{k^2 \times 2k^2 }. 
\end{align}
This exposes the fact that $H_{\parallel}$ has $k^2$ zero modes, which come from solutions to 
\begin{align}
    \delta W_{\parallel} A^{-1} + A \delta U_{\parallel} = 0,
\end{align}
and thus result from the internal symmetry. Noting that
\begin{align}
    JJ^{\top} = \Lambda_{+}^{1/2} ( AA^{\top})^{-1} \Lambda_{+}^{1/2} \otimes I_{k} + \Lambda_{+} \otimes AA^{\top}
\end{align}
is positive-definite assuming $A$ is invertible, the non-zero eigenvalues in this sector are determined by the eigenvalues of this $k^2 \times k^2$ matrix. 

Therefore, in the end, we find that the eigenvalues of $H$ split into sectors: one $k^2$-dimensional sector determined by the eigenvalues of 
\begin{align}
    \Lambda_{+}^{1/2} ( AA^{\top})^{-1} \Lambda_{+}^{1/2} \otimes I_{k} + \Lambda_{+} \otimes AA^{\top},
\end{align}
one $2k$-dimensional sector determined by the eigenvalues of 
\begin{align}
    \begin{pmatrix}
        A^{-1} \Lambda_{+} A^{-\top} & - \lambda_{j} I_{k} \\ 
        - \lambda_{j} I_{k} & \lambda_{j} A^{\top} A
    \end{pmatrix}
\end{align}
for each $j \in \{k+1,\ldots,d\}$ (giving $d-k$ sectors in all), and a remaining sector of $k^2$ zero modes. This gives, in all $k^2 + k^2 + 2k(d-k) = 2kd$ eigenvalues, as we expect. We note that, as its lower two blocks commute, we have
\begin{align}
    \det\begin{pmatrix}
        A^{-1} \Lambda_{+} A^{-\top} & - \lambda_{j} I_{k} \\ 
        - \lambda_{j} I_{k} & \lambda_{j} A^{\top} A
    \end{pmatrix}
    &= \det( \lambda_{j} A^{-1} \Lambda_{+} A^{-\top} A^{\top} A - \lambda_{j}^{2} I_{k} )
    \\
    &=   \det( \lambda_{j} A^{-1} \Lambda_{+} A - \lambda_{j}^{2} I_{k} )
    \\
    &= \lambda_{j}^{k} \prod_{i=1}^{k} (\lambda_{i} - \lambda_{j})
\end{align}
upon applying the Weinstein–Aronszajn identity in passing to the third line. This shows that the Hessian on each of these sectors is positive-definite if 
\begin{align}
    \lambda_{i} > \lambda_{j} \qquad  i \in \{1,\ldots,k\}, j \in \{k+1,\ldots,d\}.
\end{align}
We can then compute that, in general, the Hessian has trace
\begin{align}
    \tr(H) 
    &= \tr[\Lambda_{+}^{1/2} ( AA^{\top})^{-1} \Lambda_{+}^{1/2} \otimes I_{k} + \Lambda_{+} \otimes AA^{\top}] + \sum_{j=k+1}^{d} \tr\begin{pmatrix}
        A^{-1} \Lambda_{+} A^{-\top} & - \lambda_{j} I_{k} \\ 
        - \lambda_{j} I_{k} & \lambda_{j} A^{\top} A
    \end{pmatrix}
    \\
    &= d \tr[\Lambda_{+} (AA^{\top})^{-1}] + \tr(\Sigma) \tr(AA^{\top}) .
\end{align}
With the gauge-fixing $A = S^{1/2}$, 
\begin{align}
    \tr(H) = d\tr(\Lambda_{+} S^{-1}) + \tr(\Sigma)\tr(S).
\end{align}

Though the eigenvalues in general depend on $A$ in full detail, there is an important case in which they simplify. Suppose that we fix the positive gauge
\begin{align}
        A=S^{1/2},\qquad
        S=\diag(s_1,\ldots,s_k),
\end{align}
with
\begin{align}
        s_1\geq s_2\geq\cdots\geq s_k\geq 1,
        \qquad
        \lambda_1\geq\lambda_2\geq\cdots\geq \lambda_k>0.
\end{align}
matching the ordering of the eigenvalues. These assumptions will be satisfied in the effective model we derive later. Then, the non-zero eigenvalues in the within-subspace sector are
\begin{align}
        \mu_{ab}^{\parallel}
        =
        \lambda_a\left(s_b+\frac{1}{s_a}\right),
        \qquad
        a,b\in[k],
\end{align}
together with the $k^2$ zero modes coming from the internal $GL(k)$ symmetry. For each fixed $a$,
\begin{align}
        \mu_{a1}^{\parallel}
        \geq
        \mu_{a2}^{\parallel}
        \geq
        \cdots
        \geq
        \mu_{ak}^{\parallel},
\end{align}
since
\begin{align}
        \mu_{ab}^{\parallel}-\mu_{ad}^{\parallel}
        =
        \lambda_a(s_b-s_d).
\end{align}
Moreover, the diagonal is ordered:
\begin{align}
        \mu_{11}^{\parallel}
        \geq
        \mu_{22}^{\parallel}
        \geq
        \cdots
        \geq
        \mu_{kk}^{\parallel},
\end{align}
because $s\mapsto s+s^{-1}$ is increasing on $[1,\infty)$.

For the out-of-subspace sector, each pair $a\leq k<j$ gives a $2\times2$ block
\begin{align}
        H_{aj}^{\perp}
        =
        \begin{pmatrix}
        \lambda_a/s_a & -\lambda_j\\
        -\lambda_j & \lambda_j s_a
        \end{pmatrix}.
\end{align}
Following our analysis in the $k=1$ case, its eigenvalues are
\begin{align}
        \mu_{aj,\pm}
        =
        \frac12
        \left[
        \frac{\lambda_a}{s_a}
        +
        \lambda_j s_a
        \pm
        \sqrt{
        \left(
        \frac{\lambda_a}{s_a}
        +
        \lambda_j s_a
        \right)^2
        -
        4\lambda_j(\lambda_a-\lambda_j)}
        \right].
\end{align}
Since $\mu_{aj,+}\leq \tr H_{aj}^{\perp}$, and since
$\lambda_j\le \lambda_a$, we have
\begin{align}
        \mu_{aj,+}
        \leq
        \frac{\lambda_a}{s_a}
        +
        \lambda_j s_a
        \leq
        \frac{\lambda_a}{s_a}
        +
        \lambda_a s_a
        =
        \lambda_a\left(s_a+\frac1{s_a}\right)
        =
        \mu_{aa}^{\parallel}
        \leq
        \mu_{a1}^{\parallel}.
\end{align}
Thus no out-of-subspace eigenvalue exceeds the largest within-subspace eigenvalue in the corresponding retained row. Consequently,
\begin{align}
        \lambda_{\max}(H)
        =
        \max_{a\leq k}\mu_{a1}^{\parallel}
        =
        \max_{a\leq k}
        \lambda_a\left(s_1+\frac1{s_a}\right).
\end{align}
In particular,
\begin{align}
        \lambda_1\left(s_1+\frac1{s_1}\right)
        \leq
        \lambda_{\max}(H)
        \leq
        \lambda_1(s_1+1),
\end{align}
and hence, in the strongly decoder-heavy regime $s_1\gg1$,
\begin{align}
        \lambda_{\max}(H)
        =
        \lambda_1s_1+O(\lambda_1).
\end{align}

The assumptions
\begin{align}
        \lambda_1\geq\cdots\geq\lambda_k,
        \qquad
        s_1\geq\cdots\geq s_k\geq1
\end{align}
do not totally order the within-subspace eigenvalues. Indeed, for fixed $b$,
\begin{align}
        \mu_{ab}^{\parallel}-\mu_{cb}^{\parallel}
        =
        s_b(\lambda_a-\lambda_c)
        +
        \left(\frac{\lambda_a}{s_a}-\frac{\lambda_c}{s_c}\right),
\end{align}
and the second term need not have the same sign as the first. A product partial order follows under the additional ratio-ordering condition
\begin{align}
        \frac{\lambda_1}{s_1}
        \geq
        \frac{\lambda_2}{s_2}
        \geq
        \cdots
        \geq
        \frac{\lambda_k}{s_k}.
\end{align}
Under this extra assumption,
\begin{align}
        a\leq c,\quad b\leq d
        \quad\Longrightarrow\quad
        \mu_{ab}^{\parallel}\geq \mu_{cd}^{\parallel}.
\end{align}

\subsection{Hessian spectrum for a general overcomplete linear autoencoder}

Now consider an overcomplete PCA-type solution with $k>d$, for which we have $WU=I_d$. Our starting point is again 
\begin{align}
    H(\delta W,\delta U)
    &= \tr[( W \delta U + \delta W U ) \Sigma ( W \delta U + \delta W U )^{\top}]   + 2 \tr[( \delta W \delta U ) \Sigma (WU - I_{d} )^{\top}]
    \\
    &= \tr[( W \delta U + \delta W U ) \Sigma ( W \delta U + \delta W U )^{\top}] ,
\end{align}
where we note that the second term vanishes thanks to the assumption that the reconstruction is exact. We now vectorize the first term. As 
\begin{align}
     \vectorize( W \delta U + \delta W U ) &= ( I_d \otimes W) \vectorize(\delta U) + (U^{\top} \otimes I_d) \vectorize(\delta W),
\end{align}
we can write the Hessian as
\begin{align}
    H(\delta W,\delta U) = \begin{pmatrix}
         \vectorize(\delta W) \\ \vectorize(\delta U)
     \end{pmatrix}^{\top} J^{\top} J \begin{pmatrix}
         \vectorize(\delta W) \\ \vectorize(\delta U)
     \end{pmatrix}. 
\end{align}
for
\begin{align}
    J = \begin{pmatrix}
        \Sigma^{1/2} U^{\top} \otimes I_{d} & \Sigma^{1/2} \otimes W
    \end{pmatrix} . 
\end{align}
Now, as $WU = I_d$, both $W$ and $U$ must have rank $d$, hence $J$ must have at most $d^2$, a bound which is attained assuming that $\Sigma$ is full-rank. Then, noting that
\begin{align}
    J J^{\top} = \Sigma^{1/2} U^{\top} U \Sigma^{1/2} \otimes I_d + \Sigma \otimes WW^{\top}
\end{align}
is then a full-rank $d^2 \times d^2$ matrix, we find that the Hessian has $2kd-d^2$ zero modes, plus $d^2$ non-zero eigenvalues given by the eigenvalues of $JJ^{\top}$. If $\Sigma$ has zero modes, then they will generate zero modes of $JJ^{\top}$. 

If we again write the parameters at the optimum as 
\begin{align}
    W &= P^{\top} A
    \\
    U &= A^{-1} P,
\end{align}
where $P$ now satisfies $P^{\top} P = I_d$, we find that
\begin{align}
    JJ^{\top} = \Sigma^{1/2} P^{\top} (AA^{\top})^{-1} P \Sigma^{1/2} \otimes I_d + \Sigma \otimes P^\top AA^{\top} P.
\end{align}

In this case, the Hessian has the generally simple expression
\begin{align}
    \tr(H) = d \tr(\Sigma U^{\top} U) + \tr(\Sigma)\tr(WW^{\top}). 
\end{align}
If we consider a gauge in which $WW^{\top} = S$ and $U^{\top} U = S^{-1}$, then this becomes
\begin{align}
    \tr(H) = d \tr(\Sigma S^{-1}) + \tr(\Sigma)\tr(S). 
\end{align}
Further supposing that $S$ and $\Sigma$ commute, this gives
\begin{align}
    \tr(H) = d \sum_{j=1}^{d} \left(\frac{\lambda_{j}}{s_{j}} + \bar{\lambda} s_{j} \right),
\end{align}
where $\bar{\lambda} = \frac{1}{d}\tr(\Sigma)$. 

For global re-scalings of the balanced solution, with $A = \sqrt{s} I_{k}$, we have the reduction 
\begin{align}
    JJ^{\top} = (s+s^{-1}) (\Sigma \otimes I_d),
\end{align}
showing that in this case all non-zero eigenvalues are invariant under $s \mapsto 1/s$. 

\section{Detailed derivation of the effective scale dynamics for PCA}\label{app:model}

Here, we derive the effective model for scale dynamics that we presented in the main text. The accuracy of the predictions of these effective dynamics rely on two facts, which we prove below: First, that fluctuations normal to the solution manifold contract back to the manifold at a fast timescale $t \eta$. Second, that $Q_{t}$ evolves at a slow timescale $t\eta^2/B$, and on that timescale that we can neglect stochasticity in its evolution. This both justifies the effective model we present in the text, and shows that the PCA solution is locally stable. The fundamental mechanism for this is that, as we showed in Appendix~\ref{app:hessian}, the normal component of the Hessian is strictly positive-definite if the solution is a projection to the $k$-principal subspace, while otherwise it has modes with negative eigenvalue. 

In this derivation, we will assume that the batches are i.i.d., and that each is composed of $B$ i.i.d. datapoints. We assume that the data has bounded moments of all orders, \textit{i.e.}, $\mathbb{E}[ \Vert x \Vert^{2p}] < \infty$ for all $p>0$. Alternatively, we may assume that the data distribution is sub-Gaussian. In practice, we will use Gaussian data in the simulations against which we will compare the model's predictions. We will also assume that the data covariance $\mathbb{E}[xx^{\top}] = \Sigma$ is diagonal.

Consider a generic point on the solution manifold, with $(W,U) = (P^{\top} A, A^{-1} P)$. Using the polar decomposition of $A$, we gauge-fix such that $A = S^{1/2}$ for $S$ a symmetric PSD matrix. Then, $(W,U) = (P^{\top} S^{1/2}, S^{-1/2} P)$. Fix a compact subset $K$ of the cone of symmetric positive definite matrices, defined for positive constants $C>c>0$ by
\begin{align}
    K = \{ 0 \prec c I \preceq S \preceq C I\},
\end{align}
and let
\begin{align}
    \mathcal{M}_{K} = \{ (P^{\top} A, A^{-1} P)\,:\, A^{\top} A \in K\}
\end{align}
be the corresponding submanifold of the solution manifold. The assumption that $S$ is bounded will technically be required at several points in the proof.

\subsection{Fast relaxation of normal fluctuations}

We first want to control fluctuations in the normal space. We will do so via a rather abstract and generic derivation, which parallels previous analyses of how SGD behaves near solution manifolds, \textit{e.g.} by \citet{li2022what,pashakhanloo2023drift,pashakhanloo2025contribution}. 

In a small neighborhood of $\mathcal{M}_{K}$, we can decompose the full parameter vector $\theta_{t}$ as
\begin{align}
    \theta_{t} = m_{t} + \xi_{t},
\end{align}
where $m_{t} \in \mathcal{M}_{K}$ and $\xi_{t} \in N_{m_n}\mathcal{M}_{K}$ is in the normal space to $\mathcal{M}_{K}$ at $m_n$. Let $\Pi$ be the local projection onto $\mathcal{M}_{K}$, and $\Pi^{\perp}(\theta) = \theta-\Pi(\theta)$ the corresponding normal-coordinate map. Thus, $\xi_{t} = \Pi^{\perp}(\theta)$. 

Write the SGD iteration as 
\begin{align}
    \theta_{t+1} = \theta_{t} + \delta_{t},
\end{align}
where
\begin{align}
    \delta_{t} = - \eta (\nabla L(\theta_{t}) + \zeta_{t}(\theta_{t}) )
\end{align}
where $\zeta_{t}(\theta_{t})$ is the SGD noise. Thus, applying $\Pi^{\perp}$, we have
\begin{align}
    \xi_{t+1} &= \Pi^{\perp}(\theta_{t} + \delta_{t})
    \\
    &= \xi_{t} + D\Pi^{\perp}_{\theta_{t}} [\delta_{t}] + \mathcal{O}(\Vert \delta_{t} \Vert^2) ,
\end{align}
where in the second line we have made a Taylor expansion.

Now, at a given point $m$ on the manifold, the derivative of the normal-coordinate map is the projection onto the normal space:
\begin{align}
    D\Pi^{\perp}_{m} = P_{N_{m}},
\end{align}
so, assuming that $\Pi^{\perp}$ is smooth in a neighborhood around the manifold, 
\begin{align}
    D\Pi^{\perp}_{\theta_{t}} = D\Pi^{\perp}_{m_{t}+\xi_{t}} = P_{N_{m_t}} + \mathcal{O}(\Vert \xi_{t} \Vert).
\end{align}
This gives 
\begin{align}
    \xi_{t+1} = \xi_{t} + P_{N_{m_t}}\delta_{t} + \mathcal{O}(\Vert \xi_{t} \Vert\, \Vert\delta_{t}\Vert) + \mathcal{O}(\Vert \delta_{t}\Vert^2).
\end{align}

By definition, $\nabla L(m) = 0$ for all $m \in \mathcal{M}_{K}$. Then, if we Taylor expand the population loss in normal coordinates, we have
\begin{align}
    P_{N_{m}} \nabla L(m+\xi) = H_{m}^{\perp}\xi + \mathcal{O}(\Vert \xi \Vert^2),
\end{align}
where $H_{m}^{\perp}$ is the projection of the Hessian onto the normal space at $m$. Moreover, we have that $\nabla L(m_{t}+\xi_{t}) = \mathcal{O}(\Vert \xi_{t} \Vert)$. Therefore, letting
\begin{align}
    \zeta^{\perp}_{t} = P_{N_{m_t}} \zeta_{t}
\end{align}
be the projection of the SGD noise onto the normal space, we have that
\begin{align}
    P_{N_{m_t}}\delta_{t} = - \eta H_{m_{t}}^{\perp} \xi_{t} - \eta \zeta^{\perp}_{t} + \mathcal{O}(\eta\Vert \xi_{t} \Vert^2)
\end{align}
and
\begin{align}
    \Vert \delta_{t} \Vert = \mathcal{O}( \eta (\Vert \xi_{t} \Vert + \Vert \zeta_{t} \Vert) ).
\end{align}

Thus, we at last find that
\begin{align}
    \xi_{t+1} = (I-\eta H_{m_t}^{\perp})\xi_{t} - \eta \zeta_{t}^{\perp} + \epsilon_{t},
\end{align}
where the remainder term 
\begin{align}
    \epsilon_{t} = \mathcal{O}(\eta \Vert \xi_{t} \Vert^2) + \mathcal{O}(\eta \Vert \xi_{t} \Vert\, \Vert \zeta_{t} \Vert) + \mathcal{O}(\eta^{2} (\Vert \xi_{t} \Vert + \Vert \zeta_{t} \Vert)^2 )
\end{align}
has three contributions with distinct origins: the first comes from nonlinear contributions to the population gradient normal to the solution manifold, the second from the variation of the projection $P_{N_m}$ away from the manifold, and the third from the expansion of the normal-coordinate map. 

We now use this recurrence to argue that fluctuations away from the solution manifold relax rapidly. We do so supposing that the normal Hessian satisfies the uniform bound
\begin{align}
    H_{m_t}^{\perp} \succeq \mu I
\end{align}
for some $\mu>0$, which in the case of interest is true provided that the data has a spectral gap separating the $k$-principal subspace from the remaining directions (Appendix~\ref{app:hessian}). Under this assumption, the stochastic dynamics for $\xi_{t}$ are zero-reverting, and rapidly equilibrate. Given that the noise has covariance of order $1/B$, we can see that we should have 
\begin{align}
    \Vert \xi_{t} \Vert = \mathcal{O}\left(\sqrt{\frac{\eta}{B}}\right)
\end{align}
in probability. If we instead considered a solution not corresponding to the $k$-principal subspace, the Hessian would not be positive-definite, and the process would not be zero-reverting. This is the basic reason why the PCA solution manifold is locally stable under SGD, while non-principal subspace solutions are locally unstable \citep{baldi1989pca}.

To show this carefully, we fix a radius $\rho>0$, and suppose that $\xi_{t}$ lies within that tube, so that the remainder terms in $\epsilon_t$ are controlled. To be precise, let
\begin{align}
    \tau_{\rho} = \inf\{t\,:\,\Vert \xi_{t} \Vert \geq \rho\}
\end{align}
be the time of first exit from the tube. Then, using the assumption that the Hessian is uniformly positive-definite and supposing that $\eta$ is small, there exist constants $c,C>0$ independent of $\eta$ and $B$ such that
\begin{align}
    \mathbb{E}[\Vert \xi_{t+1} \Vert^2 \mathbf{1}_{t+1<\tau_{\rho}}] \leq (1-c\eta) \mathbb{E}[\Vert \xi_{t} \Vert^2 \mathbf{1}_{t+1<\tau_{\rho}}] + C \frac{\eta^2}{B},
\end{align}
where the effect of the remainder terms can be absorbed into the constants provided that the tube radius $\rho$ and learning rate $\eta$ are small enough. This assumes that $\Vert \zeta_{t}^{\perp} \Vert^4$ is uniformly bounded in expectation, which is satisfied for the datasets of interest given that $\xi_t$ lies within the tube. Iterating this recurrence, we find that 
\begin{align}
    \mathbb{E}[\Vert \xi_{t+1} \Vert^2 \mathbf{1}_{t+1<\tau_{\rho}}] \leq e^{-c\eta t} \Vert \xi_{0} \Vert^2 + C \frac{\eta}{B},
\end{align}
which means that the mean-squared normal displacement is of order $\eta/B$ after a time on the order of $\frac{1}{\eta} \log \frac{1}{\eta}$. This implies that $\Vert \xi_{t} \Vert = \mathcal{O}(\sqrt{\eta/B})$. 

We now want to argue that the probability that the exit time $\tau_\rho$ will be reached over the order-$B/\eta^2$ time interval of interest is small. Under the assumption that the data has finite moments of all orders, we can see that we can use the same argument as above to show that 
\begin{align}
    \mathbb{E}[\Vert \xi_{t+1} \Vert^{2p} \mathbf{1}_{t+1<\tau_{\rho}}] \leq  C \left( \frac{\eta}{B} \right)^{p}
\end{align}
for any $p \geq 1$, after the fast transient of order $\frac{1}{\eta} \log \frac{1}{\eta}$ steps. Let $T_{1}$ be large enough such that the transient is evaded, and $T_{2}$ be of order $B/\eta^2$, \textit{i.e.}, let $T_{2} = \lfloor \tau B/\eta^2 \rfloor$ for some order-one $\tau$. Then, by a union bound and Markov's inequality, 
\begin{align}
    \mathbb{P}\left(\max_{T_{1} \leq t \leq T_{2}} \Vert \xi_{t} \Vert > \rho\right) \leq \sum_{t=T_1}^{T_2} \mathbb{P}(\Vert \xi_{t} \Vert > \rho_{0} ) \leq C (T_{2}-T_{1}) \left(\frac{\eta}{B}\right)^{p} \leq C \eta^{p-2} B^{1-p},
\end{align}
where the constant changes between the last two inequalities, and depends on $\tau$. This goes to zero with $\eta$ for any $p>2$, hence the probability of exiting the tube tends to zero. Therefore, we can use the result above for the stopped process to show that the mean-squared displacement of $\xi_{t}$ over the interval of interest is $o(1)$. 

The important point here is that this relaxation is fast relative to the slow timescale of the evolution of $Q$. On the manifold, the mean update to $Q$ is of order $\eta^2/B$, so for an order-one change to accumulate we must wait $B/\eta^2$ steps. In contrast, we need to wait only $1/\eta$ steps for the fluctuations away from the manifold to relax back to the manifold, meaning that the two timescales have a relative separation $\eta/B$, which tends to zero as $\eta$ tends to zero. This shows that the contribution of these fluctuations to the evolution of $Q$ is $o(1)$.

\subsection{Stochastic approximation for flow along the manifold}

We now leverage the separation of timescales proved above to derive an effective flow describing the slow scale growth. This is essentially a standard exercise in stochastic approximation \citep{kushner2003stochastic}. As there is a separation of timescales, we can evaluate the evolution of $Q_{t}$ on the solution manifold and incur only an $o(1)$ error on the relevant timescale. On the manifold, we have $W_{t}^{\top} (W_{t}U_{t}-I_{d})=0$, which gives the immediate simplification 
\begin{align}
    Q_{t+1} - Q_{t} =  \eta^{2} \bigg( U_{t} \hat{\Sigma}_{t} (W_{t} U_{t} - I_{d})^{\top} (W_{t} U_{t} - I_{d}) \hat{\Sigma}_{t} U_{t}^{\top} \bigg) + o(1).
\end{align}
We now want to evaluate the mean and variance of the remaining update with respect to the batch. To do so, we decompose each datapoint as 
\begin{align}
    x = P^{\top} x_{\parallel} + x_{\perp}, 
\end{align}
where $x_{\perp}$ lies outside the principal subspace, and satisfies $Px_{\perp}=0$. This leads to the decomposition of the full data matrix as 
\begin{align}
    X_{t} = P^{\top} X_{\parallel,t} + X_{\perp,t}
\end{align}
and thus, using the fact that $W_{t} U_{t}-I_{d} = P^{\top}P - I_{d}$ and $U_{t} = S_{t}^{-1/2} P$ on the solution manifold given the gauge-fixing assumption, 
\begin{align}
    (W_{t} U_{t} - I_{d})\hat{\Sigma}_{t} U_{t}^{\top} 
    &= \frac{1}{B} (P^{\top} P - I_{d}) (P^{\top} X_{\parallel,t} + X_{\perp,t}) (X_{\parallel,t}^{\top} P + X_{\perp,t}^{\top}) P^{\top} S_{t}^{-1/2}
    \\
    &= \frac{1}{B} X_{\perp,t} X_{\parallel,t}^{\top} S_{t}^{-1/2} 
    \\
    &= \frac{1}{B} \sum_{i \in \mathcal{B}_{t}} x_{\perp,i} x_{\parallel,i}^{\top} S_{t}^{-1/2} .
\end{align}
where in the last line we introduce indexing of the datapoints, with the minibatch defined by an index set $\mathcal{B}_{t}$. Then, we have
\begin{align}
    U_{t} \hat{\Sigma}_{t} (W_{t} U_{t} - I_{d})^{\top} (W_{t} U_{t} - I_{d}) \hat{\Sigma}_{t} U_{t}^{\top}
    = \frac{1}{B^{2}} S_{t}^{-1/2}\sum_{i,j \in \mathcal{B}_{t}} (x_{\perp,i}^{\top} x_{\perp,j}) x_{\parallel,i} x_{\parallel,j}^{\top} S_{t}^{-1/2},
\end{align}
so
\begin{align}
    Q_{t+1} - Q_{t} =  \frac{\eta^2}{B^{2}} S_{t}^{-1/2}\sum_{i,j \in \mathcal{B}_{t}} (x_{\perp,i}^{\top} x_{\perp,j}) x_{\parallel,i} x_{\parallel,j}^{\top} S_{t}^{-1/2} + o(1).
\end{align}
Now, assuming that the datapoints are independent with mean zero and diagonal covariance, the mean update is 
\begin{align}
    \mathbb{E}\left[ \frac{\eta^2}{B^{2}} \sum_{i,j \in \mathcal{B}_{t}} (x_{\perp,i}^{\top} x_{\perp,j}) x_{\parallel,i} x_{\parallel,j}^{\top}  \right] = \frac{\eta^2}{B} C,
\end{align}
where we define
\begin{align}
    C = \mathbb{E}[\Vert x_{\perp} \Vert^2 x_{\parallel}x_{\parallel}^{\top}].
\end{align}
We can then write 
\begin{align}
    Q_{t+1} - Q_{t} = \frac{\eta^2}{B} S_{t}^{-1/2} C S_{t}^{-1/2} + \frac{\eta^2}{B} S_{t}^{-1/2} Z_{t} S_{t}^{-1/2} + o(1),
\end{align}
where 
\begin{align}
    Z_{t} = \frac{1}{B} \sum_{i,j \in \mathcal{B}_{t}} (x_{\perp,i}^{\top} x_{\perp,j}) x_{\parallel,i} x_{\parallel,j}^{\top} - C
\end{align}
is a noise term that is independent across timesteps. Assuming that the data has bounded eighth moments, $Z_{t}$ has bounded second moment, where the constant may depend on the batch size $B$. Thus, assuming that $S_{t}$ is bounded (which holds if the solution stays within the submanifold $\mathcal{M}_{K}$), over $T \sim B/\eta^2$ steps, we have
\begin{align}
    \var\left(\sum_{t=1}^{T} \frac{\eta^2}{B} S_{t}^{-1/2}Z_{t}S_{t}^{-1/2} \right) = \mathcal{O}\left( T \frac{\eta^4}{B^2} \right) = \mathcal{O}\left( \frac{\eta^2}{B} \right),
\end{align}
which tends to zero as $\eta \downarrow 0$ at fixed batch size. 

Combining these results, we find that if we take $\eta \downarrow 0$ while keeping $\tau = t \eta^{2}/B$ to be of order one, the discrete-time dynamics tend in probability to the solution of the corresponding limiting ODE by standard Gr\"onwall-type arguments \citep{kushner2003stochastic}. This relies on the fact that $Q(S) = S - S^{-1}$ is a Lipschitz function of $S$ provided that the solution remains within $\mathcal{M}_{K}$, \textit{i.e.}, that $S$ is uniformly bounded from below and above in L\"owner order. Assuming that this condition holds, the limiting dynamics are
\begin{align}
    \frac{d}{d\tau} (S-S^{-1}) = S^{-1/2} C S^{-1/2},
\end{align}
or, equivalently, 
\begin{align}
    \frac{dS}{d\tau} + S^{-1} \frac{dS}{d\tau} S^{-1} = S^{-1/2} C S^{-1/2}.
\end{align}

\subsection{Solution assuming co-diagonalizability}

Now, we show that the effective dynamics for $S$ can be solved analytically provided that $S$ commutes with $C$ at some reference time, which we define as $\tau=0$ without loss of generality. 

We claim that if $S$ and $C$ commute at $\tau=0$ then they do throughout time under the effective dynamics. This follows from the fact that if $S$ commutes with $C$, then so does $dS/d\tau$. To see this, we write the effective dynamics as 
\begin{align}
    L_{S}\left(\frac{dS}{d\tau}\right) = S^{-1/2} C S^{-1/2},
\end{align}
where we define the linear map 
\begin{align}
    L_{S}(X) = X + S^{-1} X S^{-1}. 
\end{align}
We can see that $L_{S}^{-1}$ is well-defined for any PSD $S$, as in vectorized form we can write
\begin{align}
    \vectorize(L_{S}(X))
    &= (I_{k} \otimes I_{k} + S^{-1} \otimes S^{-1}) \vectorize(X),
\end{align}
so we have the explicit formula
\begin{align}
    \vectorize(L_{S}^{-1}(Y)) = (I_{k} \otimes I_{k} + S^{-1} \otimes S^{-1})^{-1} \vectorize(Y).
\end{align}
This allows us to write
\begin{align}
    \frac{dS}{d\tau} = L_{S}^{-1}( S^{-1/2} C S^{-1/2}) .
\end{align}
We can now show that $L_S^{-1}(Y)$ commutes with $C$ if both $S$ and $Y$ do. If $L_{S}(X)=Y$ for $[Y,C]=0$, we have
\begin{align}
    0 = [Y,C] 
    = [L_{S}(X),C]
    = [X,C] + [S^{-1} X S^{-1},C]
    = [X,C] + S^{-1} [ X,C] S^{-1}
    = L_{S}([X,C])
\end{align}
as by assumption $S$ commutes with $C$. But, as $L_X$ is invertible, this implies that $[X,C]=0$. Therefore, we conclude the desired claim. 

As a result, if we assume that $[S(0),C]=0$, then we can choose a basis where 
\begin{align}
    S(\tau) = \diag(s_1(\tau),\ldots,s_k(\tau))
    \quad \textrm{and} \quad 
    C = \diag(\sigma_1,\cdots,\sigma_k)
\end{align}
for all times $\tau \geq 0$. Projecting into this basis, we have the decoupled equations
\begin{align}
   \frac{d s_{i}}{d\tau} + \frac{1}{s_i^2} \frac{d s_{i}}{d\tau} =  \frac{\sigma_i}{s_i} , \quad i \in [k].
\end{align}
Multiplying by $s_i$ and recognizing that
\begin{align}
    \frac{d}{d\tau} \left(\frac{1}{2} s_{i}^2 + \log s_i \right) = \left(s_{i} + \frac{1}{s_i} \right) \frac{ds_i}{d\tau},
\end{align}
we arrive at the equation
\begin{align}
    \frac{d}{d\tau} \left(\frac{1}{2} s_{i}^2 + \log s_i \right) =  \sigma_i. 
\end{align}
Integrating, we have
\begin{align}
    \frac{1}{2} s_{i}(\tau)^2 + \log s_{i}(\tau) = \frac{1}{2} s_{i}(0)^2 + \log s_{i}(0) +  \sigma_{i} \tau. 
\end{align}
We now want to solve this for $s_{i}(\tau)$. Multiplying by 2 and exponentiating, 
\begin{align}
    s_{i}(\tau)^2 e^{s_{i}(\tau)^2} = s_{i}(0)^2 e^{s_{i}(0)^2 + 2 \sigma_{i} \tau}. 
\end{align}
Now, recall that the principal branch of the Lambert $W$ function, which we will write simply as $W$, satisfies \citep{dlmf}
\begin{align}
    W(z) e^{W(z)} = z.
\end{align}
Thus, we have the solution 
\begin{align}
    s_{i}(\tau) = \sqrt{W[ s_{i}(0)^2 \exp(s_{i}(0)^2 + 2 \sigma_{i} \tau) ]},
\end{align}
as claimed in the main text. We note that $W(x) = \log x + \log \log x + o(1)$ as $x \to \infty$ \citep{dlmf}, so at long times this gives 
\begin{align}
    s_{i}(\tau) \sim \sqrt{2  \sigma_{i} \tau}. 
\end{align}
For Gaussian data with covariance eigenvalues $\lambda_1,\ldots,\lambda_k$, we have
\begin{align}
    \sigma_{i} = \left(\sum_{j=k+1}^{d} \lambda_{j} \right) \lambda_i,
\end{align}
which predicts that the growth rates along different PCs are ordered. 

We can in fact prove a stronger ordering statement, which we will use later. Suppose we start from a balanced initial condition with $s_{i}(\tau=0)=1$. Then, fixing a time $\tau$, let $s(\lambda)$ solve
\begin{align}
    s(\lambda)^2 + 2 \log s(\lambda) = 1 + a \lambda,
\end{align}
where 
\begin{align}
    a = 2 \tau \sum_{j=k+1}^{d}\lambda_{j},
\end{align}
such that $s_{i}(\tau) = s(\lambda_{i})$. We can show that 
\begin{align}
    \lambda \mapsto \frac{\lambda}{s(\lambda)}
\end{align}
is increasing in $\lambda$. By implicit differentiation, 
\begin{align}
    s'(\lambda) = \frac{a s(\lambda)}{2(s(\lambda)^2 + 1)}.
\end{align}
Then,
\begin{align}
    \frac{d}{d\lambda} \frac{\lambda}{s(\lambda)} 
    &= \frac{1}{s(\lambda)} \left[ 1 - \frac{\lambda s'(\lambda)}{s(\lambda)}\right]
    \\
    &= \frac{1}{s(\lambda)} \left[ 1 - \frac{a\lambda}{2(s(\lambda)^2+1)} \right]
    \\
    &= \frac{1}{s(\lambda)} \left[ 1 - \frac{s(\lambda)^2 + 2 \log s(\lambda)-1}{2(s(\lambda)^2+1)} \right]
\end{align}
where in the last equation we have used the equation that defines $s(\lambda)$. Therefore, $\lambda/s(\lambda)$ is increasing in $\lambda$ if
\begin{align}
    s^2 + 2 \log s - 1 < 2 (s^2 + 1),
\end{align}
or if
\begin{align}
    2 \log s < s^2 + 3 .
\end{align}
But, as $\log s < s-1$ for all $s>0$, and as $2(s-1) < s^2+3$ for all $s>0$ as it is equivalent to $2s < s^2 + 5$, we conclude that this inequality is satisfied. Therefore, for any time $\tau$, we have the ordering
\begin{align}
    \frac{\lambda_1}{s_{1}(\tau)} \geq \frac{\lambda_{2}}{s_{2}(\tau)} \geq \cdots \geq \frac{\lambda_{k}}{s_{k}(\tau)}.
\end{align}

We note also that the specialization of these results to global scale growth is easy to derive. Then, the object of interest is
\begin{align}
    [\tr(S_{t+1})-\tr(S_{t+1}^{-1})] - [\tr(S_t) - \tr(S_t^{-1})] = \Vert x_{\perp} \Vert^2 \tr[S_t^{-1} x_{\parallel} x_{\parallel}^{\top}]
\end{align}
Making the global scale \emph{Ansatz} $S = s I_k$, this leads to 
\begin{align}
    \frac{d}{d\tau} \left(s - \frac{1}{s} \right) =  \bar{\sigma},
\end{align}
where we let
\begin{align}
    \bar{\sigma} = \frac{1}{k} \tr(C). 
\end{align}
Then, it is easy to see that the same analysis as above leads to 
\begin{align}
    s(\tau) = \sqrt{W[s(0)^2 \exp(s(0)^2 + 2  \bar{\sigma} \tau)]}.
\end{align}

\subsection{Why is this a good effective model?}

We now return to the question of why this might be a good effective model for scale growth, as it is not the only way to obtain a set of effective dynamics. In particular, suppose we tried to derive an effective model directly from the dynamics of $W^{\top}W$. Expanding out the update, we have
\begin{align}
    W_{t+1}^{\top} W_{t+1}
    &= W_{t}^{\top} W_{t} - \eta (W_{t}^{\top} r_{t} x_{t}^{\top} U_{t}^{\top} + U_{t} x_{t} r_{t}^{\top} W_{t}) + \eta^{2} \Vert r_{t} \Vert^2 U_{t} x_{t} x_{t}^{\top} U_{t}^{\top}. 
\end{align}
Now, on the solution manifold, we have $W_{t}^{\top} r_{t} = 0$, and moreover that $r_{t} = - x_{t,\perp}$ and $U_{t} x_{t} = A^{-1}_{t} x_{t,\parallel}$. This leads to the simplified dynamics 
\begin{align}
    W_{t+1}^{\top} W_{t+1} - W_{t}^{\top} W_{t} = \eta^2 A^{-1} [\Vert x_{t,\perp} \Vert^2 x_{t,\parallel}x_{t,\parallel}^{\top}] A^{-\top}
\end{align}
Under the same stochastic approximation approach as before, assuming we can neglect fluctuations away from the solution manifold, this would lead to the dynamics 
\begin{align}
    \frac{dS}{d\tau} =  S^{-1/2} C S^{-1/2}
\end{align}
for $S = A^{\top} A$. This again preserves the condition that $[S(\tau),C] = 0$, meaning that we can diagonalize to obtain
\begin{align}
    s_i \frac{ds_i}{d\tau} =  \sigma_i
\end{align}
for each $i \in [k]$. As
\begin{align}
    s_i \frac{ds_i}{d\tau} = \frac{d}{d\tau} \left(\frac{1}{2} s_{i}^{2} \right),
\end{align}
this leads to
\begin{align}
    s_{i}(\tau) = \sqrt{s_{i}(0)^2 + 2  \sigma_{i} \tau}. 
\end{align}
Thus, this approach gives the same $\sqrt{2  \sigma_{i} \tau}$ growth at long times as our previous approach, but its early-time dynamics are different. 

Why is this the case, and which result should we prefer? Well, in deriving effective dynamics from the order-$\eta^2$ update from the dynamics of $Q$, we used the assumption that we are on the solution manifold only to simplify the update itself. When working with the $W^{\top} W$ dynamics directly, we used this assumption to argue that the order-$\eta$ term vanishes and also to simplify the order-$\eta^2$ term. Thus, we expect the latter effective model to incur errors as a result of fluctuations away from the solution manifold that yield non-zero values of the $\mathcal{O}(\eta)$ update; the two models are affected in similar ways by how those fluctuations change the $\mathcal{O}(\eta^2)$ term. This means that we should expect the stochastic approximation from the dynamics of $Q$ to give a better effective model. This is consistent with the fact that we see that the Lambert-$W$ solution matches experiment very closely. 

\subsection{The edge of stability for the effective model}

The effective model as derived above predicts indefinite anisotropic scale growth $s_{i}(\tau) \sim \sqrt{2\sigma_{i}\tau}$. However, this cannot be a true property of the underlying discrete-time dynamics, as the maximum eigenvalue of the Hessian of the population loss is scale-dependent, so indefinite growth would destabilize the discrete-time dynamics.\footnote{This would also violate the assumption of boundedness of $S$ that we used in the derivation of the effective model, but that technical issue can be resolved by proving the result over fixed time intervals with a bound that grows in time. The destabilization of the SGD dynamics is a more substantial violation of the assumptions. } For an eigenvalue $\mu_{i}(H)$ of the Hessian, we recall that the general edge of stability criterion is that $\eta \mu_{i}(H) \approx 2$, with the threshold being likely somewhat less than 2 for SGD \citep{cohen2021gradient,damian2023self,andreyev2025stochasticstability,emmanouilidis2026edge}.

Recall from our analysis in Appendix~\ref{app:hessian} that the within-subspace sector Hessian eigenvalues are
\begin{align}
    \mu_{ab}^{\parallel} = \lambda_{a} \left(s_{b} + \frac{1}{s_a}\right), \qquad
        a,b\in[k],
\end{align}
and that we expect these to be the dangerous modes. We have shown above that the scales satisfy the ordering
\begin{align}
    \frac{\lambda_{i}}{s_i} \geq \frac{\lambda_j}{s_j}
\end{align}
if $i < j$, if we start from a balanced initial condition. Thus, given the ordering results from Appendix~\ref{app:hessian}, we conclude that for any $a \leq c$ and $b \leq d$, we have
\begin{align}
    \mu_{ab}^{\parallel} \geq \mu_{cd}^{\parallel} .
\end{align}
For a given $b$, we then have that $\mu_{ab}^{\parallel} \geq \mu_{cb}^{\parallel}$ whenever $a \leq c$, which shows that $\mu_{1b}^{\parallel}$ is the dangerous mode for each $b$. 

Given the ordering of the scales $s_i$, we expect them to sequentially hit the stability threshold. Write the threshold condition as $\eta \lambda_{i}(H) = \kappa$ for some $0<\kappa \leq 2$. First, $s_1$ will hit a threshold at $s_1=s_1^*$ determined by
\begin{align}
    \eta\lambda_{1}\left(s_{1}^* + \frac{1}{s_1^*}\right) = \kappa . 
\end{align}
Subsequently, the lower modes will hit thresholds determined by
\begin{align}
    \eta \lambda_{1} \left(\frac{1}{s_{1}^*}+ s_{j}^* \right) = \kappa, \quad j > 1,
\end{align}
This leads to a common ceiling for all scale directions. To show this very explicitly, let $\tilde{\kappa} = \kappa/\eta\lambda_{1}$ be the normalized stability threshold, which we assume to be large as we are working at small learning rate. Then, the above conditions state that the scales satisfy
\begin{align}
    s_{1}^* = \tilde{\kappa} - \frac{1}{s_1^*} \quad \textrm{and} \quad s_{j}^* = \tilde{\kappa} - \frac{1}{s_1^*}, \quad j>1. 
\end{align}
From this, we can see immediately that we must have all $s_{j}^* = s_{1}^*$ at the threshold, where by solving the equation for $s_{1}^*$ we have
\begin{align}
    s^*_{j} = s_1^* = \frac{\tilde{\kappa} + \sqrt{\tilde{\kappa}^2-4}}{2} = \tilde{\kappa} + \mathcal{O}(\tilde{\kappa}^{-1}) ,
\end{align}
where the asymptotic expansion holds for $\tilde{\kappa} \gg 1$.

\section{Batch sharpness decreases with scale}\label{app:bs}

\citet{andreyev2025stochasticstability} proposed that networks trained with minibatch SGD operate near the Edge of Stochastic Stability, defined by when the so-called ``batch sharpness'' saturates near $2/\eta$. The batch sharpness is defined by 
\begin{align}
    \mathrm{BS} = \mathbb{E}_{\mathcal{B}}\left[ \frac{\nabla \hat{L}_{\mathcal{B}}^{\top} \hat{H}_{\mathcal{B}} \nabla\hat{L}_{\mathcal{B}}}{\Vert \nabla \hat{L}_{\mathcal{B}} \Vert^2}\right]
\end{align}
where $\hat{L}_{\mathcal{B}}$ is the loss on minibatch $\mathcal{B}$, and $\hat{H}_{\mathcal{B}}$ is its Hessian. 

In the present setting, consider a minibatch $X_{\mathcal{B}} \in \mathbb{R}^{d \times B}$ with empirical covariance $\hat{\Sigma}_{\mathcal{B}} = \frac{1}{B} X_{\mathcal{B}}X_{\mathcal{B}}^{\top}$. Then, we have
\begin{align}
    \hat{L}_{\mathcal{B}} = \frac{1}{2} \tr[(WU-I_{d}) \hat{\Sigma}_{\mathcal{B}} (WU-I_{d})^{\top}], 
\end{align}
so the minibatch gradients are
\begin{align}
    \nabla_{W}\hat{L}_{\mathcal{B}} &= (W U - I_{d}) \hat{\Sigma}_{\mathcal{B}} U^{\top} \\ 
    \nabla_{U}\hat{L}_{\mathcal{B}} &= W^{\top} (W U - I_{d}) \hat{\Sigma}_{\mathcal{B}} .
\end{align}
Using the general result from Appendix~\ref{app:hessian}, the minibatch Hessian is compactly expressed as a bilinear form acting on a perturbation $(\delta W,\delta U)$:
\begin{align}
\begin{split}
    \hat{H}_{\mathcal{B}}(\delta W,\delta U)
    &= \tr[( W \delta U + \delta W U ) \hat{\Sigma}_{\mathcal{B}} ( W \delta U + \delta W U )^{\top}] \\&\quad  + 2 \tr[( \delta W \delta U ) \hat{\Sigma}_{\mathcal{B}} (WU - I_{d} )^{\top}]. 
\end{split}
\end{align}

We now compute the batch sharpness at a PCA-type solution. Recall that such a solution satisfies
\begin{align}
    (W,U) = (P^{\top} A, A^{-1} P)
\end{align}
where $P \in \mathbb{R}^{k \times d}$ is an orthogonal projection ($PP^{\top} = I_{k}$) and $A \in \mathrm{GL}(k)$. Then, 
\begin{align}
    W U - I_{d} = P^{\top} P - I_{d} = - P_{\perp}
\end{align}
is a projection, and moreover one that is annihilated by $P$. This shows that $\nabla_{U} \hat{L}_{\mathcal{B}}=0$, while $\nabla_{W}\hat{L}_{\mathcal{B}} = - P_{\perp} \hat{\Sigma}_{\mathcal{B}} P^{\top} A^{-\top}$. As before, introduce coordinates for the data matrix 
\begin{align}
    X_{\mathcal{B}} = P^{\top} X_{\parallel,\mathcal{B}} + X_{\perp,\mathcal{B}},
\end{align}
such that
\begin{align}
    P X_{\mathcal{B}} = X_{\parallel,\mathcal{B}} \quad \textrm{and} \quad P_{\perp} X_{\mathcal{B}} = X_{\perp,\mathcal{B}}.
\end{align}
Then, 
\begin{align}
    \Vert \nabla \hat{L}_{\mathcal{B}} \Vert^2 
    &= \Vert \nabla_{W}\hat{L}_{\mathcal{B}} \Vert^2 
    = \frac{1}{B^2} \Vert P_{\perp} X_{\mathcal{B}} X_{\mathcal{B}}^{\top} P^{\top} A^{-\top} \Vert^2 
    =  \frac{1}{B^2} \tr(K_{\mathcal{B}} H_{\mathcal{B}}).
\end{align}
where we have defined the $B \times B$ matrices
\begin{align}
    K_{\mathcal{B}} = X_{\parallel,\mathcal{B}}^{\top} (AA^{\top})^{-1} X_{\parallel,\mathcal{B}}
    \quad \textrm{and} \quad H_{\mathcal{B}} = X_{\perp,\mathcal{B}}^{\top} X_{\perp,\mathcal{B}} .
\end{align}
Moreover, 
\begin{align}
    \nabla_{W}\hat{L}_{\mathcal{B}} U = - P_{\perp} \hat{\Sigma}_{\mathcal{B}} P^{\top} (AA^{\top})^{-1} P,
\end{align}
hence
\begin{align}
    \hat{H}_{\mathcal{B}}(\nabla_{W}\hat{L}_{\mathcal{B}},\nabla_{U}\hat{L}_{\mathcal{B}})
    &= \tr[( \nabla_{W}\hat{L}_{\mathcal{B}} U ) \hat{\Sigma}_{\mathcal{B}} ( \nabla_{W}\hat{L}_{\mathcal{B}} U )^{\top}] 
    = \frac{1}{B^3} \tr[K_{\mathcal{B}}^{2} H_{\mathcal{B}}].
\end{align}
Therefore, we can write the batch sharpness on the PCA solution manifold as
\begin{align}
    \mathrm{BS} = \mathbb{E}_{\mathcal{B}}\left[ \frac{1}{B} \frac{\tr[K_{\mathcal{B}}^{2} H_{\mathcal{B}}]}{\tr(K_{\mathcal{B}} H_{\mathcal{B}})}\right] . 
\end{align}
Under a global re-scaling $AA^{\top} \mapsto s AA^{\top}$, we have $K_{\mathcal{B}} \mapsto s^{-1} K_{\mathcal{B}}$, and thus
\begin{align}
    \mathrm{BS} \mapsto \frac{1}{s} \mathrm{BS}.
\end{align}
Therefore, on the PCA solution manifold the batch sharpness is \emph{decreasing} in global scale; it is in fact precisely homogeneous of degree $-1$. 

This property is shared by what \citet{andreyev2025stochasticstability} define as the modified batch sharpness, which amounts to separately averaging the numerator and denominator:
\begin{align}
    \mathrm{MBS} = \frac{\mathbb{E}_{\mathcal{B}}[ \nabla \hat{L}_{\mathcal{B}}^{\top} \hat{H}_{\mathcal{B}} \nabla\hat{L}_{\mathcal{B}}]}{\mathbb{E}_{\mathcal{B}}[ \Vert \nabla \hat{L}_{\mathcal{B}} \Vert^2]}
    = \frac{1}{B} \frac{\mathbb{E}_{\mathcal{B}}[\tr(K_{\mathcal{B}}^{2} H_{\mathcal{B}})]}{\mathbb{E}_{\mathcal{B}}[\tr(K_{\mathcal{B}}  H_{\mathcal{B}})]} .
\end{align}
The advantage of considering this quantity is that it can be computed analytically for the Gaussian data model on which we focus. As in Appendix~\ref{app:model}, work in a basis where the data covariance $\Sigma$ and the scale matrix $S$ are jointly diagonal. Then, $X_{\parallel,\mathcal{B}}$ and $X_{\perp,\mathcal{B}}$ are independent, and thus $H_{\mathcal{B}}$ is independent of $K_{\mathcal{B}}$. Then, using the fact that $\mathbb{E}_{\mathcal{B}}[H_{\mathcal{B}}] = (\sum_{j=k+1}^{d}\lambda_{j}) I_{B}$, we find that
\begin{align}
    \mathrm{MBS} = \frac{1}{B} \frac{\mathbb{E}_{\mathcal{B}}[\tr(K_{\mathcal{B}}^{2}) ]}{\mathbb{E}_{\mathcal{B}}[\tr(K_{\mathcal{B}}  )]} ,
\end{align}
as the overall constant of proportionality cancels. Using Isserlis' theorem for Gaussian moments, we find that
\begin{align}
    \mathbb{E}_{\mathcal{B}}[\tr(K_{\mathcal{B}})] &= B a_{1},
    \\
    \mathbb{E}_{\mathcal{B}}[\tr(K_{\mathcal{B}}^2)] &= B a_{1}^2 + (B+1) a_{2}
\end{align}
where we let
\begin{align}
    a_{1} = \sum_{i=1}^{k} \frac{\lambda_i}{s_i}
    \quad \textrm{and} \quad 
    a_{2} = \sum_{i=1}^{k} \frac{\lambda_i^2}{s_i^2}. 
\end{align}
This gives
\begin{align}
    \mathrm{MBS} = \frac{1}{B} \left(a_{1} + (B+1) \frac{a_2}{a_1}\right).
\end{align}

Before the edge of stability is reached, we can use the solution of the effective model to study the time evolution of the modified batch sharpness. In fact, we can show that, if we start from a balanced initial condition, 
\begin{align}
    \frac{d}{d\tau} \mathrm{MBS} < 0
\end{align}
along the flow. To show this, we recall from Appendix~\ref{app:model} that, assuming Gaussian data, the dynamics under the effective model are 
\begin{align}
    \frac{ds_i}{d\tau} = \sigma_{\perp} \lambda_{i} \frac{s_i}{s_{i}^2 + 1} , \quad s_{i}(0)=1
\end{align}
assuming a balanced initial condition, where $\sigma_{\perp} = \sum_{j=k+1}^{d} \lambda_{j}$ is the variance outside the principal subspace. Moreover, we recall that the explicit solution is given implicitly by 
\begin{align}
    s_{i}^{2} + 2 \log s_{i} - 1 = 2 \sigma_{\perp} \lambda_{i} \tau. 
\end{align}
Then, we see immediately that 
\begin{align}
    \frac{d}{d\tau} \log s_{i} > 0,
\end{align}
so 
\begin{align}
    \frac{d}{d\tau} a_{1} 
    &= \sum_{i=1}^{k} \frac{d}{d\tau} \frac{\lambda_{i}}{s_{i}} = -\sum_{i=1}^{k} \frac{\lambda_{i}}{s_{i}} \frac{d}{d\tau} \log s_{i} < 0.
\end{align}
What requires more care is to show that $a_2/a_1$ is decreasing. Expanding out the derivatives, we can see that we need to show that 
\begin{align}
    - a_{1}^2 \frac{d}{d\tau} \frac{a_2}{a_1} 
    &= 2 a_{1} \sum_{i=1}^{k} \frac{\lambda_i^2}{s_i^2} \frac{d}{d\tau}\log s_i - a_{2} \sum_{i=1}^{k} \frac{\lambda_i}{s_i} \frac{d}{d\tau} \log s_i 
    \\
    &= \sum_{i=1}^{k} \frac{\lambda_{i}^3}{s_{i}^{3}} \frac{d}{d\tau} \log s_{i} + \sum_{i<j} \frac{\lambda_{i}\lambda_{j}^2}{s_{i}s_{j}^2} F_{ij}
\end{align}
is positive, where we have defined
\begin{align}
    F_{ij} = \left( 2 \frac{\lambda_{i}}{s_{i}} \frac{s_{j}}{\lambda_{j}} -  1 \right) \frac{d}{d\tau} \log s_{i} + \left( 2  - \frac{\lambda_{i}}{s_{i}} \frac{s_{j}}{\lambda_{j}}\right) \frac{d}{d\tau} \log s_{j} .
\end{align}
As suggested by our definition, our remaining task is to show that $F_{ij}>0$, as the first sum is clearly positive. Focus on some entry with $i<j$. We know from Appendix~\ref{app:model} that if $i \leq j$ then
\begin{align}
    \frac{\lambda_{i}}{s_i} \geq \frac{\lambda_{j}}{s_{j}}.
\end{align}
Therefore, 
\begin{align}
    \alpha = \frac{\lambda_{i}}{s_i} \frac{s_{j}}{\lambda_{j}} \geq 1 , 
\end{align}
in terms of which 
\begin{align}
    F_{ij} = (2 \alpha - 1 ) \frac{d}{d\tau} \log s_{i} + (2-\alpha) \frac{d}{d\tau} \log s_{j} . 
\end{align}
If $1 \leq \alpha \leq 2$, then both coefficients $2\alpha-1$ and $2-\alpha$ are non-negative, and at least one is strictly positive, so clearly $F_{ij} > 0$. If $\alpha > 2$, then more work is required. Let 
\begin{align}
    r_{i} = \frac{d}{d\tau} \log s_{i}.
\end{align}
We claim that if $i < j$, then $r_{i} > \frac{1}{2}r_{j}$; we will prove this below. Given this claim, for any $\alpha\geq 1$ we have
\begin{align}
    F_{ij} = (2\alpha-1) r_{i} + (2-\alpha) r_{j} > (2\alpha-1) \frac{1}{2} r_{j} + (2-\alpha) r_{j} = \frac{3}{2} r_{j} > 0.
\end{align}
To prove the claim that if $i < j$, then $r_{i} > \frac{1}{2}r_{j}$, we start by letting $y_{i} = s_{i}^{2}$, and recall from before that if $i<j$ then $y_{i} \geq y_{j} \geq 1$. The dynamics of $s_{i}$ imply that
\begin{align}
    r_{i} = \sigma_{\perp}\lambda_i \frac{1}{y_{i}+1}. 
\end{align}
At the initial condition $\tau = 0$, all of the scales $y_{i} = 1$, so we have $r_{i}(\tau=0) = \frac{1}{2}\sigma_{\perp} \lambda_{i}$, meaning that $r_{i} \geq r_{j} \geq 0$ for any $i<j$ given the eigenvalues are ordered, hence the claim follows trivially. For $\tau > 0$, all $y_{j}>1$, and we can use the explicit solution $s_{i}^{2} + \log s_{i}^2 - 1 = 2 \sigma_{\perp} \lambda_{i} \tau$ to write $\sigma_{\perp}\lambda_{i}$ in terms of the solution $y_{i}$, which leads to 
\begin{align}
    r_{i} = \frac{1}{2\tau} \phi(y_{i}) \quad \textrm{for} \quad \phi(y) = \frac{y+\log y - 1}{y+1}. 
\end{align}

We are then left with a self-contained problem about the scalar function $\phi(y) = \frac{y+\log y - 1}{y+1}$, defined for $y \geq 1$: we must prove that if $x \geq y > 1$, then $\phi(x) > \frac{1}{2} \phi(y)$. By inspection, $\phi(y=1) = 0$, and moreover $\phi(y) \geq 0$ for all $y \in [1,\infty)$. Differentiating, we see that
\begin{align}
    \phi'(y) = \frac{3+y^{-1}-\log(y)}{(1+y)^2}
\end{align}
has a unique zero at $y_{\star}=1/W(e^{-3})$, the unique solution to $3+y^{-1}-\log(y)=0$. From this, we find that $\phi$ is strictly increasing on $[1,y_{\star}]$ and strictly decreasing on $[y_{\star},\infty)$. An elementary lower bound on $y_{\star}$ is that $y_{\star} > e^{3}$, which holds because $\phi'(e^{3}) = \frac{e^{-3}}{(e^3+1)^2}>0$. At the unique global maximum $y_{\star}$, we have $\phi(y_{\star}) = 1 + W(e^{-3}) \leq 1 + W(1) < 2$, and from this we see that $\phi(y) < 2$ for all $y$. This upper bound also can be directly proved by noting that it is equivalent to the statement that $y + \log y - 1 < 2y + 2$, which in turn is equivalent to the obvious inequality $\log y < y + 3$. 

We can now divide the proof of the desired claim into two cases. If $\phi(x) \geq \phi(y)$, then we can immediately conclude that $\phi(x) > \frac{1}{2} \phi(y)$ because we know that $\phi(y) \geq 0$ for any $y \geq 1$. If instead $\phi(x) < \phi(y)$ for $x \geq y > 1$, then we must have $x > y_{\star}$. Now, as noted before, we know that $\phi(y) < 2$, so the claim follows if $\phi(x) \geq 1$. We can see that $\phi(x) \geq 1$ is equivalent to the inequality $x + \log x - 1 \geq x+1$, which is in turn equivalent to $\log x \geq 2$, or $x \geq e^{2}$. But, we know that $x > y_{\star}$, and from before that $y_{\star} > e^{3}$, which is greater than $e^{2}$. As a result, we conclude that $x > e^{2}$ and thus that $\phi(x) \geq 1$, and thus the desired result. This concludes the proof.

\section{Nonlinear autoencoders}\label{app:nonlinear}

As we focus on linear autoencoders, a natural question is whether any of the phenomenology generalizes to nonlinear models. 

\subsection{Extending the PCA manifold to homogeneous nonlinear models}

Consider a general nonlinear autoencoder 
\begin{align}
    f_{\theta,\varphi}(x) = g_{\theta}(h_{\varphi}(x)),
\end{align}
where $h_{\varphi}:\mathbb{R}^{d} \to \mathbb{R}^{k}$ and $g_{\theta}:\mathbb{R}^{k} \to \mathbb{R}^{d}$ are the encoder and decoder, with parameter vectors $\varphi$ and $\theta$, respectively. We will usually suppress the subscripts, and write simply $f$, $g$, and $h$. Suppose that the model has a scale symmetry, in the sense that its output $f_{\theta,\varphi}(x)$ is invariant for any example $x$ under transformations of the form
\begin{align} \label{eqn:scale_symmetry}
    (\theta,\varphi) \mapsto (a\theta,a^{-1}\varphi), \quad a>0.
\end{align}
If we train the model to minimize an $L^2$ reconstruction loss $L = \frac{1}{2} \mathbb{E}_{x}[ \Vert f(x) - x \Vert^2]$ via gradient flow $\partial_{t} \theta = - \nabla_{\theta} L$, $\partial_{t} \varphi = - \nabla_{\varphi} L$, then it is well-known that the scale symmetry \eqref{eqn:scale_symmetry} means that the corresponding Noether charge 
\begin{align} \label{eqn:q_definition}
    q = \Vert \theta \Vert^2 - \Vert \varphi \Vert^2
\end{align}
is conserved, \textit{i.e.}, $\partial_{t} q = 0$ \citep{du2018balanced,kunin2021neural,tanaka2021noether}. To see this, note that our assumption that $f(x)=g(h(x))$ is invariant under transformations $(\theta,\varphi) \mapsto (a\theta,a^{-1}\varphi)$ for $a>0$ implies that the per-example loss $\ell(x) = \frac{1}{2}\Vert f(x)-x\Vert^2$ is invariant. This implies that
\begin{align}\label{eqn:gradient_difference}
    0 = \frac{d}{da} \ell(x)\bigg|_{a=1} = \langle \nabla_{\theta}\ell(x),\theta\rangle - \langle\nabla_{\varphi}\ell(x),\varphi\rangle, 
\end{align}
which in turn implies that
\begin{align}
    \partial_{t} q &= 2 (\langle \theta,\partial_{t}\theta\rangle -  \langle \varphi,\partial_{t}\varphi\rangle) \\&= -2 \mathbb{E}[\langle\theta, \nabla_{\theta}\ell(x)\rangle - \langle\varphi,\nabla_{\varphi}\ell(x)\rangle] 
    \\
    &= 0.
\end{align}

Under SGD updates $\theta_{t+1} = \theta_{t} - \eta \nabla_{\theta} \hat{L}_{t}$, $\varphi_{t+1} = \varphi_{t} - \eta \nabla_{\varphi} \hat{L}_{t}$ for $\hat{L}_{t} = \mathbb{E}_{x\sim\mathcal{B}_t}[\ell(x)]$, we have
\begin{align}
    q_{t+1} 
    &= q_{t} - 2 \eta \mathbb{E}_{x\sim\mathcal{B}_t}[\langle\theta, \nabla_{\theta}\ell(x)\rangle - \langle\varphi,\nabla_{\varphi}\ell(x)] 
    + \eta^{2} (\Vert \nabla_{\theta} \hat{L}_{t} \Vert^2 - \Vert \nabla_{\varphi} \hat{L}_{t} \Vert^2)
    \\
    &= q_{t} + \eta^{2} (\Vert \nabla_{\theta} \hat{L}_{t} \Vert^2 - \Vert \nabla_{\varphi} \hat{L}_{t} \Vert^2),\label{eqn:general_scale_iteration}
\end{align}
where the order-$\eta$ term vanishes identically by \eqref{eqn:gradient_difference}. 

We now observe that we can partially extend the PCA solution geometry to these more general nonlinear models. Consider a particular example $x$. Let its latent encoding be $z = h(x)$, and the residual be $r = g(z)-x$, such that the reconstruction error is $\ell(x)=\frac{1}{2} \Vert r \Vert^2$. By the chain rule, the parameter gradients appearing in the update \eqref{eqn:general_scale_iteration} for $q_{t}$ are $\nabla_{\theta}\ell = (\partial g/\partial\theta)^{\top}r$ and $\nabla_{\varphi}\ell = (\partial h/\partial\varphi)^{\top} (\partial g/\partial z)^{\top} r$. For an overcomplete autoencoder, perfect reconstruction is possible, leading to $r=0$ and $q_{t+1}=q_{t}$. In contrast, for an undercomplete autoencoder with $k<d$, it is not generally possible to make the residual vanish for an arbitrary example. 

Geometrically, we can then decompose the residual $r$ into components normal and tangential to the decoder manifold $\mathcal{D} = \{g(z)\,:\,z\in\image h\}$, which we write as $r_{\perp}$ and $r_{\parallel}$, respectively. By definition, $\left({\partial g}/{\partial z} \right)^{\top} r_{\perp}=0$. Thus, the encoder gradient $\nabla_{\varphi}\ell$ is sensitive only to $r_{\parallel}$, while the decoder gradient $\nabla_{\theta}\ell$ is sensitive to both $r_{\parallel}$ and $r_{\perp}$. Assuming that $k \ll d$, it is possible for the undercomplete autoencoder to make the $k$-dimensional tangential component $r_{\parallel}$ quite small. As a result, as the decoder gradient has an additional contribution from $r_{\perp}$ relative to the encoder gradient, we expect undercomplete autoencoders to generically favor decoder-heavy solutions so long as $\Vert r_{\parallel} \Vert \ll \Vert r_{\perp} \Vert$. In the extreme, we can consider what we might refer to as \emph{projection-type} solutions that have $r_{\parallel}=0$; these are in some sense an analog of the PCA solutions in the linear case. On the manifold of such solutions, the imbalance dynamics reduce to 
\begin{align}
    q_{t+1} = q_{t} + \eta^{2} \Vert \nabla_{\theta} \hat{L}_{t} \Vert^2 ,
\end{align}
that is, there is a systematic drift towards larger decoder scales. 

\subsection{Per-neuron scale symmetries in ReLU autoencoders}

In the section above we considered the case in which the $\mathrm{GL}(k)$ symmetry of a linear autoencoder is reduced all the way down to a single $\mathbb{R}_{>0}$. However, for a standard ReLU autoencoder
\begin{align}
    f(x) = W \sigma(U x),
\end{align}
where $\sigma(t) = \max(0,t)$ acts elementwise, has a larger symmetry group defined by re-scaling each hidden neuron, in addition to permutation symmetries. The continuous component of the symmetry group is simply the neuron-wise rescalings $\mathbb{R}_{>0}^{k}$, which correspond to $(W,U) \mapsto (W A, A^{-1} U)$ for $A = \diag(a_1,\ldots,a_k)$ a diagonal matrix with positive diagonal elements $a_{i} > 0$ \citep{du2018balanced,tanaka2021noether}.

A more tractable expression follows by letting $w_{i}$ be the $i$-th column of $W$, and $u_{i}^{\top}$ be the $i$-th row of $U$. Then, 
\begin{align}
    f(x) = \sum_{i=1}^{k} w_{i} \sigma( u_{i}^{\top} x)
\end{align}
and the action of the symmetry group is that, for each $i$, $(w_{i},u_{i}) \mapsto (a_{i} w_{i}, a_{i}^{-1} u_{i})$. This formulation also makes it obvious that there are $k$ Noether charges corresponding to this symmetry:
\begin{align}
    q_{i} = \Vert w_{i} \Vert^2 - \Vert u_{i} \Vert^2. 
\end{align}
The charge related to global rescaling is then simply the sum of these per-neuron charges: $q = \sum_{i=1}^{k} q_{i}$. The proof that these charges are individually conserved under gradient flow parallels the proof for homogeneous rescalings considered above: for any $i \in [k]$, 
\begin{align}
    0 = \frac{\partial}{\partial a_i} \ell(x) \bigg|_{A=I_k}
    = \langle \nabla_{w_i} \ell(x), w_{i}\rangle - \langle \nabla_{u_i} \ell(x), u_{i} \rangle ,
\end{align}
so 
\begin{align}
    \partial_{t} q_{i} = 2 (\langle w_{i}, \partial_{t} w_{i}\rangle - \langle u_{i}, \partial_{t} u_{i} \rangle )
    = 2 \mathbb{E}[\langle \nabla_{w_i} \ell(x), w_{i}\rangle - \langle \nabla_{u_i} \ell(x), u_{i} \rangle]
    = 0.
\end{align}
Under SGD, we have the general update
\begin{align}
    q_{t+1,i} - q_{t,i} = \eta^{2} ( \Vert \nabla_{w_i}\hat{L}_{t} \Vert^2 - \Vert \nabla_{u_i} \hat{L}_{t} \Vert^2).
\end{align}
We show in Figure~\ref{fig:relu-scale-imbalance} that, empirically, SGD training leads to decoder-heavy imbalance for these networks.

\section{Experimental details} \label{app:experiments}

\begin{figure}[h]
    \centering
    \includegraphics[width=0.6\linewidth]{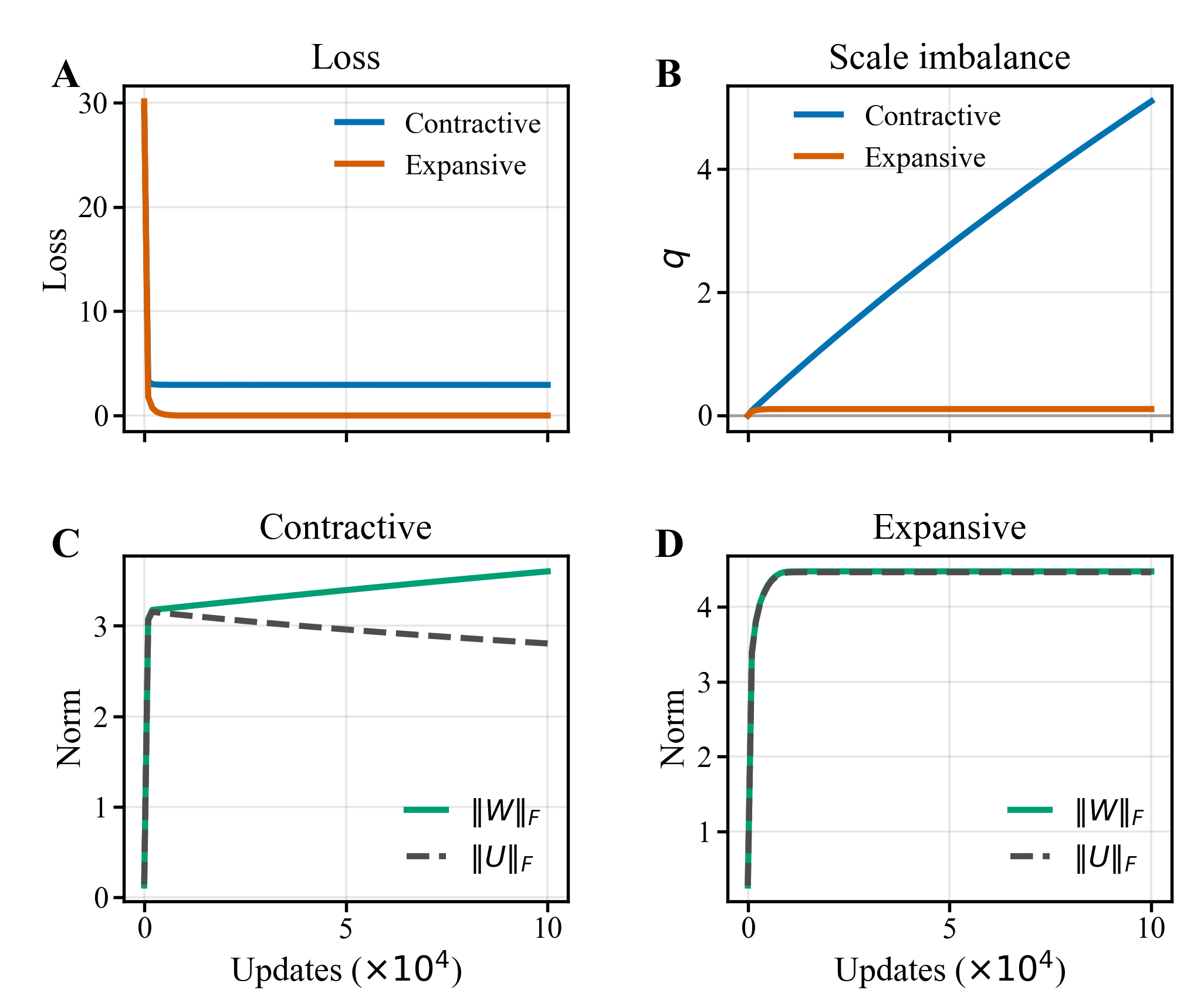}
    \caption{
    Continual minibatch training drives scale imbalance in linear autoencoders.
    (A) Validation loss for contractive and expansive linear autoencoders during online training with random initialization at the onset.
    (B) Scale imbalance is measured by
    $q = \Vert W\Vert_{F}^{2} - \Vert U \Vert_{F}^{2}$. 
    (C,D) Frobenius norms of decoder and encoder weights,
    $\|W\|_F$ and $\|U\|_F$, for contractive and expansive autoencoders, respectively (see \S\ref{app:experiments} for experimental details).}
    \label{fig:linear-scale-imbalance}
\end{figure}

\begin{figure}[h]
    \centering
    \includegraphics[width=0.6\linewidth]{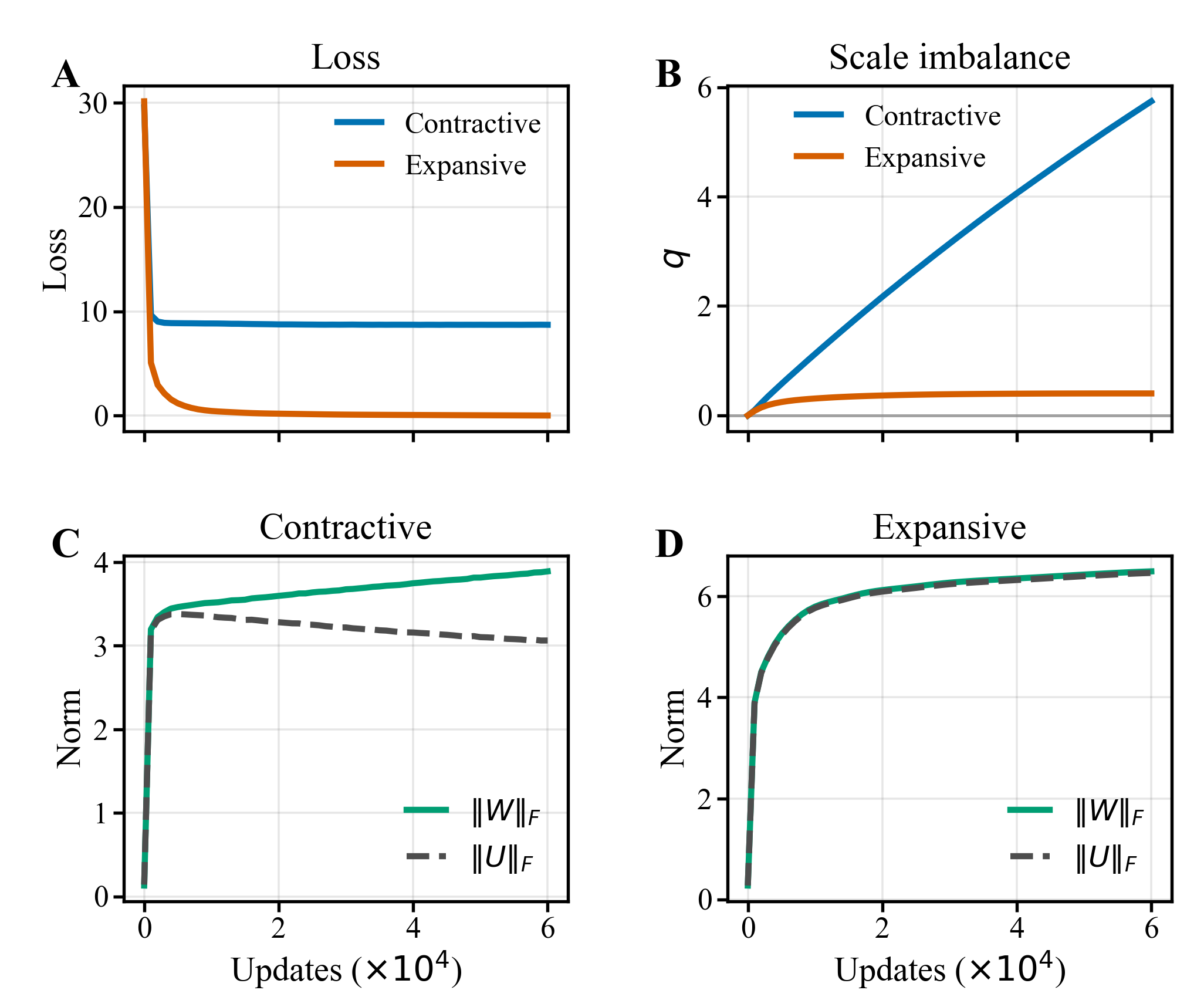}
    \caption{
    Continual minibatch training drives scale imbalance in homogeneous autoencoders.
    (A) Validation loss for contractive and expansive ReLU autoencoders during online training with random initialization at the onset.
    (B) Scale imbalance is measured by
    $q$ as defined in Eq.~\eqref{eqn:q_definition}. In this case we have a decoder weight matrix $W$ and an encoder weight matrix $U$, so $q = \Vert W\Vert_{F}^{2} - \Vert U \Vert_{F}^{2}$. 
    (C,D) Frobenius norms of decoder and encoder weights,
    $\|W\|_F$ and $\|U\|_F$, for expansive and contractive ReLU autoencoders, respectively (see \S\ref{app:experiments} for experimental details). }
    \label{fig:relu-scale-imbalance}
\end{figure}

The simulations were performed with two-layer autoencoders of the form $\hat{x} = W \phi(Ux)$,
with zero bias, input dimension \(n=20\), and either a contractive hidden layer
\(k=10<n\) or an expansive hidden layer \(k=40>n\). The data were generated
online from a diagonal Gaussian distribution,
$x \sim \mathcal{N}(0,\Sigma)$, where
$\Sigma = \mathrm{diag}(\lambda_1,\ldots,\lambda_n)$
is the covariance eigenvalues were geometrically spaced from
\(\lambda_1=12\) to \(\lambda_n=0.18\) (concretely, $\lambda_i = \lambda_1(\frac{\lambda_n}{\lambda_1})^\frac{i-1}{n-1}$). Training used minibatch SGD with batch
size \(B\), learning rate \(\eta\), mean-squared
reconstruction loss, and fresh Gaussian samples at every update. 
Unless specified, all networks had linear activation $\phi(z_i)=z_i$.
The specific parameters for each plot are as follows.
\begin{itemize}
\item 
Figure~\ref{fig2:workshop}A/\!\! Figure~\ref{fig:linear-scale-imbalance}: $n=20$, $k=10$ (contractive) and $k=40$ (expansive), $B=48$, $\eta=0.003$.
\item 
Figure~\ref{fig:relu-scale-imbalance}: $n=20$, $k=10$ (contractive) and $k=40$ (expansive), $B=48$, $\eta=0.003$. $\phi(z_i) = \operatorname{max}(0,z_i)$ (ReLU activation).
\item 
Figure~\ref{fig2:workshop}B:  $n=20$, $k=10$, variable $B$ and $\eta$, as shown on the plot.
\item 
Figure~\ref{fig2:workshop}C: $n=20$, $k=10$, $B=32$, $\eta = 0.005$.
\item 
Figure~\ref{fig2:workshop}D-E: $n=20$, $k=10$, $B=128$, $\eta=0.02$.
\item 
Figure~\ref{fig2:workshop}F: $n=20$, $k=10$, $B=128$, $\eta \in \{0.014, 0.016, \ldots, 0.026\}$.
\end{itemize}
Networks in Figure~\ref{fig2:workshop}A/\!\! Figure~\ref{fig:linear-scale-imbalance} and Figure~\ref{fig:relu-scale-imbalance}
were initialized with i.i.d. zero-mean Gaussian weights with standard deviation \(0.01\), while the contractive networks in the other plots were initialized at the PCA solution with the identity scale. Specifically, for those networks,
\[
W_0 =
\begin{bmatrix}
I_k \\
0_{(n-k)\times k}
\end{bmatrix},
\qquad
U_0 =
\begin{bmatrix}
I_k & 0_{k\times(n-k)}
\end{bmatrix},
\qquad
W_0U_0 =
\operatorname{diag}
\left(
I_k,0_{(n-k)\times(n-k)}
\right),
\]
resulting in an initial scale matrix $S_0 = W_0^{\top}W_0 = I_k$.



\clearpage 

\end{document}